\documentclass{article}

\usepackage{arxiv}
\usepackage[utf8]{inputenc} 
\usepackage[T1]{fontenc}    
\usepackage[breaklinks, colorlinks=true, citecolor=black, urlcolor=black, linkcolor=black]{hyperref}
\usepackage{url}            
\usepackage{booktabs}       
\usepackage{amsfonts}       
\usepackage{nicefrac}       
\usepackage{microtype}      
\usepackage{graphicx}
\usepackage{natbib}
\usepackage{doi}
\usepackage{amsmath,amssymb,amsthm,mathrsfs,bbm,amsfonts,bm}
\usepackage{color,graphicx,cases,caption,enumerate}
\usepackage{float,multicol}
\usepackage[dvipsnames]{xcolor}
\usepackage{colortbl}
\usepackage{booktabs}
\usepackage{multirow}
\usepackage{setspace}
\usepackage{capt-of}
\usepackage{array}
\usepackage[title]{appendix}
\newcolumntype{C}[1]{>{\centering\let\newline\\\arraybackslash\hspace{0pt}}m{#1}}
\usepackage{natbib}
\usepackage{placeins} 
\usepackage[breaklinks, colorlinks=true, citecolor=black, urlcolor=black, linkcolor=black]{hyperref}
\usepackage{subcaption}
\usepackage{tikz}
\usetikzlibrary{fit,positioning, arrows.meta}
\usepackage{makecell}

\newcommand{\p}{\ensuremath{\mathbb{P}}}

\newcommand{\e}{\ensuremath{\mathbb{E}}}

\title{Modeling Spatial Extremal Dependence of Precipitation Using Distributional Neural Networks}

\author{{Christopher Bülte}\thanks{Current affiliation: Ludwig-Maximilians-Universität, Munich} \\
	Chair of Statistical Methods and Econometrics\\
	Karlsruhe Institute of Technology (KIT)\\
	Karlsruhe, Germany \\
	\texttt{buelte@math.lmu.de} \\
	\And
	{Lisa Leimenstoll} \\
	Chair of Statistical Methods and Econometrics\\
	Karlsruhe Institute of Technology (KIT)\\
	Karlsruhe, Germany \\
	\texttt{lisa.leimenstoll@kit.edu} \\
 	\And
	{Melanie Schienle} \\
	Chair of Statistical Methods and Econometrics\\
	Karlsruhe Institute of Technology (KIT)\\
	Karlsruhe, Germany \\
	\texttt{melanie.schienle@kit.edu} \\
}

\renewcommand{\headeright}{}
\renewcommand{\undertitle}{}
\renewcommand{\headerleft}{Modeling Spatial Extremal Dependence of Precipitation Using Distributional Neural Networks}
\renewcommand{\shorttitle}{}

\begin{document}
\maketitle

\begin{abstract}
\noindent
In this work, we propose a simulation-based estimation approach using generative neural networks to determine dependencies of precipitation maxima and their underlying uncertainty in time and space.
Within the common framework of max-stable processes for extremes under temporal and spatial dependence, our methodology allows estimating the process parameters and their respective uncertainty, but also delivers an explicit nonparametric estimate of the spatial dependence through the pairwise extremal coefficient function. We illustrate the effectiveness and robustness of our approach in a thorough finite sample study where we obtain good performance in complex settings for which closed-form likelihood estimation becomes intractable.
With the proposed technique we study the spatial dependence of two of the most severe extreme precipitation events of the last decades in Germany: The Ahr valley flood in 2021 and the Elbe flood in 2002, which both had a massive, deadly impact. Our methodology allows for assessing the dependence structure per event, revealing significant structural differences in the different flood events and areas. Beyond the considered setting, the presented methodology and its main generative ideas also have great potential for other applications.
\end{abstract}


\section{Introduction}
Recent years have been marked by an increase in the number, magnitude, and spatial concentration of extreme precipitation events that affected businesses, infrastructure, and public health and safety. In fact, empirical evidence shows that the last decades have been characterized by a high incidence of flooding events \citep{bloschl2020current} and a generally elevated risk of flooding in Europe \citep{mitchell2003european}. Due to heavy rainfall, several flood disasters have occurred across Europe and Germany - most notably the July 2021 flood in central Europe, which counts among the five most costly disasters in Europe in the last half century and claimed more than 180 lives in Germany alone mostly in the Ahr region \citep{Mohr.2023, Bosseler.2021} and the August 2002 Elbe flood disaster, which at the time caused the largest flood-related losses ever recorded in Germany and was widely regarded as a millennial flood with a return period of less than 1000 years \citep{mechler2003disaster}. 
As extreme precipitation is the main driver of flood risk, understanding these extreme events and their underlying spatio-temporal occurrence structure is crucial. For assessing extremes of geophysical systems, the key statistical workhorse models are max-stable processes \citep{Davison.2012b}. Under standard regularity conditions, these processes arise as the limit of pointwise maxima of random fields and allow for a flexible but parsimonious modeling of the underlying dependence structure accounting for the scarcity of observations in extremes. Despite the simple parametric form of max-stable processes, however, the precision and often the general feasibility of estimation and prediction still suffers from the challenging set-up of extremes  where by design the available effective number of observations is small relative to the large dimension of the vector of precipitation measurement instances across space. Often, this requires simplifying assumptions on the dependence structure that are not only hard to justify but might miss out on essential points in practice.

In this work, we propose a general simulation-based estimation approach for extreme precipitation events based on generative neural networks for parameters of max-stable processes and their corresponding spatial dependence. For this, our methodology pretrains distributional neural networks on data from simulated max-stable processes before these neural networks are used to estimate model parameters for observational data. Our technique allows for a fully general dependence structure and goes beyond pure point estimates, providing the full predictive distribution and thus quantifying the uncertainty of parameters of max-stable processes. Moreover, as a direct result of the proposed technique, we also obtain a nonparametric estimate of the distribution of the spatial dependence between any two points as measured by the pairwise extremal coefficient function.  
We also illustrate that our approach is robust to the specific type of max-stable model chosen in the training step. Our work builds on recent efforts in neural networks for parameter estimation \citep{Lenzi.2023b, SainsburyDale.2022} with advances in training generative neural networks with proper scoring rules for probabilistic predictions \citep{pacchiardi2022likelihoodfree, Chen.2022, buelte2025}. An extensive simulation study shows excellent finite sample performance of the proposed method, investigating in particular robustness to misspecification of the specific type of max-stable model employed for the simulation part.
We apply the proposed technique to annual precipitation maxima in particularly exposed regions, focusing on two major events in Germany, covering the Ahr region flood catastrophe in western Germany in July 2021 and the extreme precipitation event that caused the Elbe floods in eastern Germany in 2002. We compare their dependence structure in order to investigate whether the spatial extremal dependence differs between these two regions during flood events.
The new estimator directly reveals the spatial dependence structure of precipitation extremes and allows for a comparison of  its strength and shape over time. We find that during the time of extreme flooding events in plane rather than valley areas, this dependence is much more pronounced than during regular years, even when accounting for uncertainty. This suggests that the provided technique might help to detect dangerous flood situations.

In our approach, we rely on max-stable processes that model spatial extremes via pointwise block maxima that are especially tailored to strong time series dependencies. Max-stable processes \citep{Davison.2012b} are widely applied, for example, for spatially modeling wind gusts \citep{Ribatet.2013}, durations of extreme rainfall \citep{https://doi.org/10.1002/2017WR022231}, or analyzing yearly maximum precipitation \citep{reich2012hierarchical}. Besides the max-stable models, there exist the conventional threshold-based approaches that also build on classical extreme value theory. A common approach is to model precipitation extremes with the peaks over threshold (POT) method, which considers extremes as events that exceed a specified threshold. Here, the focus is less on the time-dependence structure but more on a large-scale spatial dimension \citep[see, e.g.][]{halmstad2013analysis, Wadsworth.2022} with the recent exception of \cite{vandeskog2024fast}. In all of the model cases, estimation is difficult due to the complexity of the corresponding likelihoods and the small effective sample sizes for extremes. In particular, for max-stable processes, a closed-form likelihood function and estimator are generally not computationally feasible. The most common approach to circumvent that issue is to consider a composite likelihood method, by replacing the full likelihood with a pairwise likelihood \citep{Padoan.2010}. While this has been successful in estimating model parameters, it comes with a loss of statistical efficiency \citep{Huser.2013, Castruccio.2016} and neglects high-order dependencies. Several improvements have been suggested, such as the Vecchia approximation \citep{Huser.2022b}, incorporating occurrence times of maxima \citep{Stephenson.2005} or using expectation-maximization \citep{Huser.2019} that modify the set of assumptions for estimation but still require rather strong conditions to hold.

In contrast to the classical statistical methods, different approaches have been proposed that circumvent the usage of the likelihood function entirely. Usually, such techniques are simulation-based, augmenting the scarce data for extremes with simulated data points. 
The most popular method is the approximate Bayesian computation (ABC) framework \citep{beaumont.2002, Franks.2020}, which retrieves a posterior parameter distribution by comparing selected summary statistics of observations and simulations via a suitable loss function. Although the ABC method has been directly applied to specific models, such as max-stable processes \citep{Erhardt.2012,fearhnead.2021}, the choice of summary statistic and loss function is not straightforward and requires careful calibration. In addition, the approach requires a large number of simulations to generate a reliable estimation, making it computationally demanding. In a recent work, \cite{vandeskog2024fast} introduce a method for fast simulation of precipitation extremes, highlighting the importance of simulations. More recently, there has been a focus on novel methods for parameter estimation and likelihood-free inference, mainly by employing neural networks trained on simulated processes. Based on the ABC approach, \cite{Creel.2017} proposes to train a neural network on an informative summary statistic and apply their method to two different econometric models. Similarly, \cite{Rai.2023} estimate parameters of the generalized extreme value distribution by training a neural network on a summary statistic based on extreme quantiles. Their results show similar accuracy, but a reduced computation time, compared to maximum likelihood estimation. Concerning spatial data \cite{Gerber.2021} estimate the local covariance structure of Gaussian processes via convolutional neural networks. Utilizing a similar method, \cite{Lenzi.2023b} directly estimate the parameters of max-stable processes. \cite{SainsburyDale.2022} propose the so-called neural Bayes estimator, which trains a neural network by minimizing the Bayes risk and which they apply to different spatial models, including max-stable processes. The above-mentioned neural-network-based approaches share the advantage that they require only small observational sample sizes and tend to be significantly faster than classical methods. However, the methods so far fall short of providing calibrated uncertainty estimates and typically require the underlying form of the model to be known.

The remainder of this article is organized as follows. The precipitation data is presented in Section~\ref{sec:data}. Section~\ref{sec:method} outlines the theory regarding max-stable processes, as well as the theoretical background of our approach. Section~\ref{sec:simulation} entails the specific implementation of the neural network, evaluation metrics and results of simulation studies, as well as additional robustness checks. In Section~\ref{sec:application} we apply our method to the described precipitation data to model extreme precipitation in two different regions of Germany.
A final discussion is given in Section~\ref{sec:discussion}.

\section{Data}
\label{sec:data}
We consider historical data of daily precipitation maxima for two specific regions in Germany, where two of the most severe floodings of the last decades occurred. The data is provided by the German National Meteorological Service DWD \citep{hyras} and based on measurements from 1300 stations across different countries, which are regridded to a resolution of $1 \times 1\ \text{km}^2$. The data is freely available from an online archive\footnote[1]{\url{https://opendata.dwd.de/climate_environment/CDC/grids_germany/daily/hyras_de/precipitation/}}, dating back to January 1, 1930.
\begin{figure}[htb]
 \centering
\includegraphics[width=\linewidth]{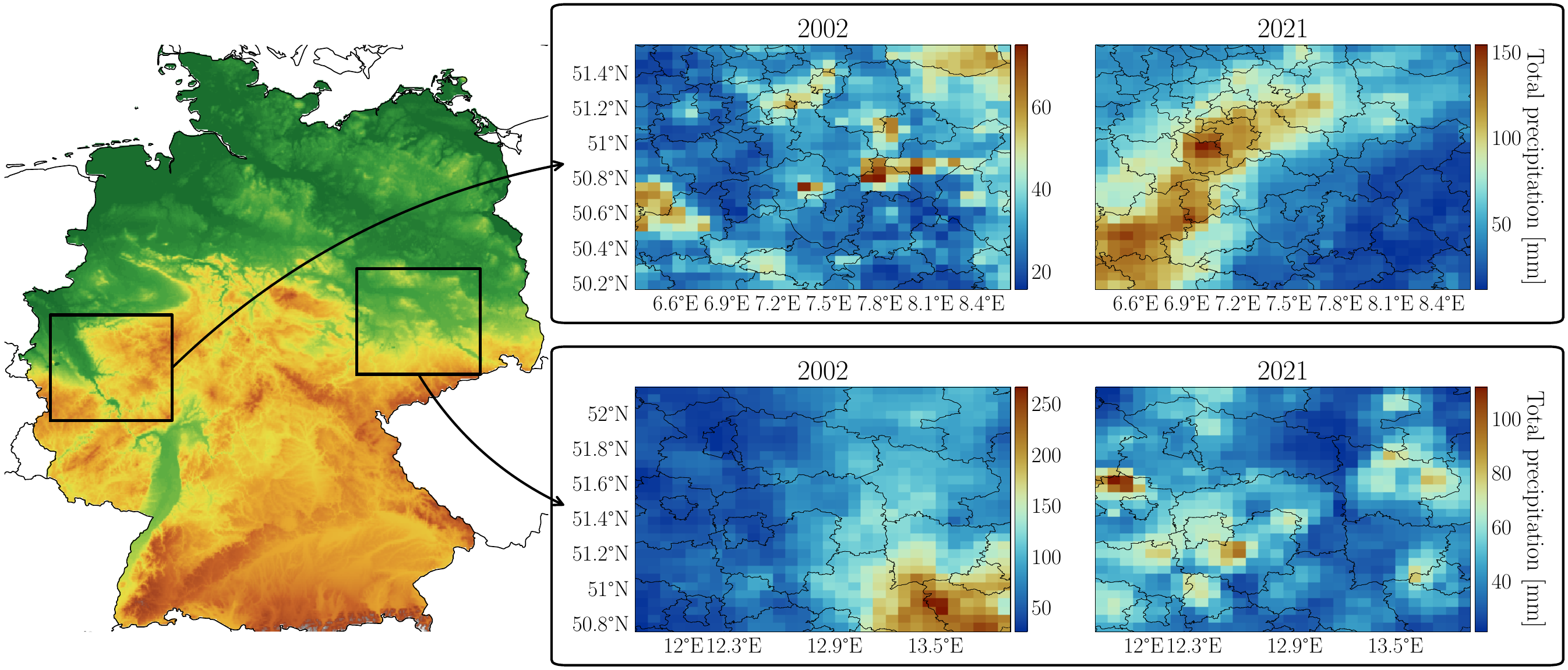}
\caption[Monthly precipitation maxima]{The figure shows the two selected domains with maximum yearly precipitation in mm, aggregated over the three summer months, as well as their geographic location. The Ahr region (upper) with the flood event in 2021 and the Elbe region (lower) with the flood event in 2002}
\label{fig:yearly_tp}
\end{figure}

We use the data for modeling extreme monthly precipitation over the two selected areas highlighted in \autoref{fig:yearly_tp}, each transformed to a $30 \times 30$ grid of around 150km $\times$ 150km, covering an area of $22500\ \text{km}^2$. Similar to \cite{Oesting.2022}, we aggregate the three summer months, June, July, and August, to obtain a total of 96 yearly precipitation maxima for the entire sample of completed years 1930-2025.

The first area covers the Ahr region, where in July 2021 an extremely heavy precipitation event took place, with more than 150 mm precipitation over an extensive area within approximately 15 to 18 hours. The resulting floods, mainly concerning the Ahr valley, led to at least 180 fatalities, 40.000 people affected and an estimated damage of around EUR 33 billion\footnote{Munich Re, Hurricanes, cold waves, tornadoes: Weather disasters in USA dominate natural disaster losses in 2021, \emph{Press report} 10.01.2022 (\href{https://www.munichre.com/en/company/media-relations/media-information-and-corporate-news/media-information/2022/natural-disaster-losses-2021.html}{link}, accessed on 07.11.2023)}. For an overview and a description of the event, see, for example \cite{Bosseler.2021} or \cite{Mohr.2023}. 

The second area covers the region around the federal state of Saxony in eastern Germany, which was the area most severely impacted by the 2002 Elbe flood. During the ten-day period from August 1 to 10, around 60 mm of precipitation fell across the Elbe River drainage basin area, followed by a phase of particularly intense rainfall between August 11 and 13 that drastically compounded the flooding issues. Overall, the event caused damages totaling  around EUR 9.1 billion for Germany alone. For more details on the event and its consequences, compare \cite{mechler2003disaster}, \cite{habersack2005analyse} or \cite{muller2010hochwasserrisikomanagement}.

\section{Methodology}
\label{sec:method}
\subsection{Max-stable processes}
\label{ssec:setup}

As we are analyzing aggregated precipitation maxima (i.e. block maxima)  a natural modeling framework is that of max-stable processes. Under standard regularity assumptions these processes constitute the unique non-degenerate limit of renormalized pointwise block maxima of i.i.d. random fields making them a widely used tool for analyzing spatial extremes \citep{Davison.2012b, Davison.2015}. Following \cite{Schlather.2002}, a max-stable process is given as
\begin{equation}
\label{eq:max-process-representation}
    \boldsymbol{Z}(\boldsymbol{x}) = \max_{i \geq 1} \xi_i \boldsymbol{Y}_i(\boldsymbol{x}), \quad \boldsymbol{x} \in \mathcal{X},
\end{equation}
where $\mathcal{X} \subseteq \mathbb{R}^d$, $\{\xi_i, \ i \in \mathbb N \}$ denote points of a nonnegative Poisson process on $(0, \infty)$, $\boldsymbol{Y}(\cdot)$ is a nonnegative stochastic process defined on $\mathbb{R}^d$ such that $\e [\boldsymbol{Y}(\boldsymbol{x})] = 1, \ \forall \boldsymbol{x} \in \mathbb{R}^d$ and $\boldsymbol{Y}_i(\cdot)$ are i.i.d. copies of $\boldsymbol{Y}(\cdot)$. The process in (\ref{eq:max-process-representation}) is normalized to unit Fréchet margins, i.e. $\p (\boldsymbol{Z}(\boldsymbol{x})\leq z) = \exp(-1 / z), \ \forall \boldsymbol{x} \in \mathcal{X},\  z> 0$. Different choices of $\boldsymbol{Y}(\cdot)$ lead to different max-stable processes. We consider two main popular model classes: The Brown-Resnick \citep{Kabluchko.2009} and the Schlather \citep{Schlather.2002} model.

The Brown-Resnick model sets $\boldsymbol{Y}_i(\bm x) = \exp \{ \boldsymbol{\epsilon}_i(\bm x) - \gamma(\boldsymbol{h})\}$ in (\ref{eq:max-process-representation}), where $\boldsymbol{\epsilon}_i$ are independent copies of a centered Gaussian process with (semi-) variogram $\gamma(\boldsymbol{h})$ and spatial separation $\boldsymbol{h}$.
Due to their flexibility, Brown-Resnick models are often applied in practice \citep[compare][]{Thibaud.2016, Oesting.2017}. Here, we consider an isotropic model, where the corresponding correlation or variogram function only depends on the distance $h = \|\bm x_1 - \bm x_2 \|_2$ and is typically specified in terms of the range and smoothness parameters $\lambda \in \mathbb{R}_+$ and $\nu \in \mathbb{R}_+$. 
A prominent special case of the Brown-Resnick model is the Smith model \citep{Smith.1990} that holds if $\bm \epsilon_i(\bm x) = \bm x^\top \Sigma^{-1} X$ and $\ X \sim \mathcal{N}(\bm 0, \Sigma)$.

The Schlather model is obtained from representation (\ref{eq:max-process-representation}) by setting $\boldsymbol{Y}_i(\bm x) = \sqrt{2\pi} \max \{0, \boldsymbol{\epsilon}_i(\bm x) \}$, where $\boldsymbol{\epsilon}_i(\bm x)$ are i.i.d. copies of a standard Gaussian process with correlation function $\rho(\boldsymbol{h})$, where $\bm h =  \bm x_2 - \bm x_1$.
The Schlather model leads to stationary max-stable processes that have been successfully used, for example, to model temperature maxima \citep{Davison.2012} or minima \citep{Erhardt.2012}. 
 The correlation function is usually chosen from a choice of valid parametric families that only depend on the Euclidean distance $h = \|\bm x_2 - \bm x_1 \|_2$, and therefore leads to isotropic processes. The most common subtypes are 

\begin{itemize}
    \item {Powered Exponential} with
    $
        \rho(h) = \exp\left( - \left( \frac{h}{\lambda} \right)^{\nu} \right),
        \quad \lambda > 0,\; 0 < \nu \le 2
    $

    \item {Whittle–Matérn} with
    $
        \rho(h) = \frac{2^{1-\nu}}{\Gamma(\nu)} \left( \frac{h}{\lambda} \right)^{\nu} K_{\nu}\left( \frac{h}{\lambda} \right),
        \quad \lambda > 0,\; \nu > 0
    $

    \item {Cauchy} with
    $
        \rho(h) = \left( 1 + \left( \frac{h}{\lambda} \right)^2 \right)^{-\nu},
        \quad \lambda > 0,\; \nu > 0
    $
\end{itemize}

where \( \lambda, \nu \) are the so-called range and smoothness parameter of the correlation function, \( \Gamma \) is the gamma function and \( K_{\nu} \) is the modified Bessel function of the third kind with order \( \nu \).

For the max-stable process $Z(\boldsymbol{x})$ the joint cumulative distribution at any finite collection of spatial sites $\{\boldsymbol{x}_1, \ldots, \boldsymbol{x}_k \} \subset \mathcal{X}$ is given by $\p(Z_1(x_1)\leq z_1,\ldots,Z_k(x_k)\leq z_k)=\exp(-V(z_1,\ldots,z_k))$, where $V(z_1,\ldots,z_k)$ denotes the exponent measure \citep{Haan.2006}, encoding the dependence structure in the extremes. Moreover for any $z>0$ $\p(Z_1(x_1)\leq z,\ldots,Z_k(x_k) \leq z)= \exp(-\theta/z)$, where $\theta= V(1,\ldots,1) \in [1,k]$ is known as the extremal coefficient of $Z$. The corresponding probability density function of $\boldsymbol{Z}(\boldsymbol{x})$ can be derived from (\ref{eq:max-process-representation}) as
\begin{equation}
    \label{eq:full_likelihood}
    f(z_1, \ldots, z_k; \bm{\gamma}) = \exp\left(-V(z_1, \ldots, z_k)\right) \sum_{\pi \in \mathcal{P}_k} (-1)^{|\pi |} \prod_{j=1}^{|\pi|} V_{\pi_j}(z_1, \ldots, z_k),
\end{equation}
where $\boldsymbol{\gamma}=(\lambda,\nu)^T$ is the parameter vector, $\mathcal{P}_k$ denotes the set of all partitions $\{ \pi_1, \ldots, \pi_p \}$ of the set $\{\bm x_1, \ldots, \bm x_k\}$ and $|\pi | = p$ is the size of the partition $\pi$ and $V_{\pi_j} = \frac{\partial^{|\pi_j|}}{\partial z_{\pi_j}}V(z_1, \ldots, z_k)$ is partial derivative of $V(z_1, \ldots, z_k)$ with respect to the variables indexed by the set $\pi_j$. Even if $V$ is available in closed form, the number of terms involved in \eqref{eq:full_likelihood} quickly explodes, as it is summed over the set of all possible partitions and it has been shown that the expression is not computationally tractable for $k>12$ \citep{Castruccio.2016}. A typical workaround is to consider the pairwise likelihood \citep{Padoan.2010, Davis.2013}, which is defined as
\begin{equation}
    \label{eq:pairwise_likelihood}
    \ell_p (\bm{\gamma}; \bm{z}) = \sum_{i=1}^{k-1} \sum_{j=i+1}^k w_{i,j} \log f(z_i, z_j;\bm{\gamma}),
\end{equation}
where $\bm{z} = (z_1, \ldots, z_k)$ is a single observation and $f(\cdot, \cdot;\bm \theta)$ is the bivariate pdf, obtained from \eqref{eq:full_likelihood}. The weights $w_{i,j}$ are typically chosen based on a cutoff distance \citep{Padoan.2010}. The estimator can be shown to be consistent and asymptotically normal under standard assumptions, but it is not asymptotically efficient, meaning that its asymptotic variance exceeds the minimal attainable variance and can therefore be relatively large. 

Similar to the extremal coefficient, we measure and analyze the dependence structure across spatial extremes by the so-called pairwise extremal coefficient function, defined by
\begin{equation}
    \label{eq:pairwise_extremal_coefficient}
    \theta( h) = - z \log \p (\boldsymbol{Z}(\bm x_1) \leq z, \boldsymbol{Z}(\bm x_2) \leq z) = \e [ \max \left\{ \boldsymbol{Y}(\bm x_1), \boldsymbol{Y}(\bm x_2) \right\}],
\end{equation}
where $\boldsymbol{Y}(\cdot)$ and $\boldsymbol{Z}(\cdot)$ are as in \eqref{eq:max-process-representation}. Since all models considered are isotropic, the pairwise extremal coefficient function only depends on the spatial distance $h = \| \boldsymbol{x}_1 - \boldsymbol{x}_2 \|_2$ of any two points $\bm x_1$ and $\bm x_2$.  By definition $\theta( h)$ is directly related to the probability that two spatial sites do not exceed a common threshold $z$ and therefore provides a measure of spatial dependence. Note that $\theta(h)$ only takes values in the range $[1,2]$, with the lower bound corresponding to complete dependence and the upper bound to independence of the two spatial locations. While the pairwise extremal coefficient function is analytically available for a wide range of models, in practice, it is often estimated using the so-called F-madogram \citep{Cooley.2006} based on the following relation from the max-stability property of  $\bm Z$ \citep{Cooley.2006}
\begin{equation}
\label{eq:f_madogram}
    \theta(h) = \frac{1+2\nu_F(h)}{1-2\nu_F(h)}
\end{equation}
where $\nu_F(h) := \frac{1}{2}\mathbb{E}\left[\left| F(\bm Z(\bm x_1)) - F(\bm Z(\bm x_2)) \right| \right]$ is the so-called F-madogram. In $\nu_F(h)$, the function $F$ denotes the cumulative distribution function of $\bm Z(\bm x)$. For an estimator of $\theta(h)$ in \eqref{eq:f_madogram} a simple empirical estimator $\hat{\nu}_F$ of $\nu_F(h)$ based on rank statistics \citep{Ribatet.2013} is used as a plug-in. Any estimator of the F-madogram requires a sufficient amount of observations over time for a valid empirical approximation of the expectation in $\nu_F(h)$.

\subsection{Estimation framework}
\label{ssec:parameter_estimation}

\begin{figure}
\centering
\begin{tikzpicture}
\tikzstyle{connection}=[ultra thick,every node/.style={sloped,allow upside down},draw=black,opacity=1]
\node[draw = black, inner sep = 8pt, thick, rounded corners] (gamma) at (0,0,0) {$\boldsymbol{\gamma} \sim \Pi$};
\node[draw = black, inner sep = 10pt, thick, rounded corners,fill=yellow!80!black, fill opacity=0.2,text opacity=1, above right = 1.5cm and 1cm] (process) at (gamma.east) { Max-stable model};
\node[draw = black, rounded corners, inner sep = 8pt, thick, right = 1cm] (z_s) at (process.east) { $\bm Z(\bm s ; \bm \gamma)$};
\node[draw = black, inner sep = 10pt, thick,  rounded corners,fill=yellow!80!black, fill opacity=0.2,text opacity=1, below = 0.93cm ] (train) at (z_s.south) { Train networks ${F}^1_\phi$ and $ F^2_{\psi}$};
\node[draw = black, inner sep = 10pt, thick, rounded corners, fill=yellow!80!black, fill opacity=0.2,text opacity=1, right = 1cm] (pred) at (train.east) { Predict};
\node[draw = black, inner sep = 10pt, thick, rounded corners, fill=yellow!80!black, fill opacity=0.2,text opacity=1, above = 1cm, align = center] (data) at (pred.north) { Observational\\  data};
\node[draw = black, rounded corners, inner sep = 8pt, thick, above right = 0.8cm and 1cm] (param_pred) at (pred.east) { $\left(\hat{\bm \gamma}_j\right)_{j=1}^m$};
\node[draw = black, rounded corners, inner sep = 8pt, thick, right= 1cm] (theta_pred) at (pred.east) { $\big(\hat{\bm \theta}_j(h_{\Delta})\big)_{j=1}^m$};
\draw[-{Stealth[length=3mm]},shorten <=9pt, shorten >=9pt, thick] (gamma.east) -- (process.west);
\draw[-{Stealth[length=3mm]},shorten <=5pt, shorten >=5pt, thick] (process.east) -- (z_s.west);
\draw[-{Stealth[length=3mm]},shorten <=5pt, shorten >=5pt, thick] (z_s.south) -- (train.north);
\draw[-{Stealth[length=3mm]},shorten <=9pt, shorten >=9pt, thick] (gamma.east) -- (train.west);
\draw[-{Stealth[length=3mm]},shorten <=5pt, shorten >=5pt, thick] (train.east) -- (pred.west);
\draw[-{Stealth[length=3mm]},shorten <=5pt, shorten >=5pt, thick] (data.south) -- (pred.north);
\draw[dashed,-{Stealth[length=3mm]},shorten <=9pt, shorten >=9pt, thick] (pred.east) -- (param_pred.west);
\draw[dashed,-{Stealth[length=3mm]},shorten <=5pt, shorten >=5pt, thick] (pred.east) -- (theta_pred.west);
\end{tikzpicture}
\caption{Flowchart visualizing our procedure for estimating spatial dependence. Max-stable processes are simulated according to a chosen prior distribution and a specified model, such as Brown-Resnick. The neural networks $\mathcal{F}^1_\phi$ and $\mathcal{F}^2_\psi$ are then trained via the corresponding energy score. For available observations, the model predicts discrete probabilistic estimates of the parameter and the pairwise extremal coefficient function on a pre-specified grid.} 
\label{fig:method_workflow}
\end{figure}

In this section, we present the proposed estimation framework that works in two main steps, comprising a training and an evaluation step. In the training step, we generate simulated data from max-stable processes in order to train neural networks targeting the model parameters and the dependence function. The trained networks are then used on observational data to obtain predictions of the parameters of interest. The suggested procedure does not require the specification or minimization of a likelihood function and thus avoids the caveats outlined above that require restrictive distributional assumptions for feasibility in practice. Please see the schematic overview of the procedure in Figure \ref{fig:method_workflow}. The details of the training and prediction step are described below.

In substep one of the training, we simulate $n$-times from a specific max-stable (exponential) model on an equally spaced (regular) grid of $k$ spatial locations uniformly distributed on a domain $\mathcal{D}$. 
Similar to \cite{Erhardt.2012} in each of the $n$ data generation rounds, we draw the defining $\gamma$ parameter in the above models from an uninformative uniform prior $\Pi \sim \mathcal{U}(\bm{a,b})$ and obtain for each $\gamma(i)$ a field of $k$ simulated data points on the spatial grid, which we write as the $k$-vector $\mathbf{Z}^{[i]}(\gamma(i))$. For the choice of the tuning parameters $a$ and $b$ we refer to the following subsection where we present a simple data-driven heuristic.

In substep two of the training, we use the generated $\boldsymbol{Z}=(\mathbf{Z}^{[1]}(\gamma(1)), \ldots,\mathbf{Z}^{[n]}(\gamma(n)))$ to train a first neural network $F^1_\phi$ that outputs a posterior distribution $\mathbb{Q}_{\phi}(\cdot \mid \boldsymbol{Z}^{[i]})$ for $\gamma(i)$ for each $i \in \{1,\ldots, n\}$. Moreover, with $\mathbf{Z}$ we also train a second network $F^2_\psi$ that targets the pairwise extremal coefficient function $\theta$ on a grid $h_{\Delta}=\{h_1,\ldots,h_{l}\}$ with $ l\in \mathbb{N}$ of pre-specified support points of spatial radii yielding $\bm \theta(h_{\Delta})=(\theta(h_1),\ldots ,\theta(h_l))\in [1,2]^l$, with $\theta(h_i)<\theta(h_j)$ for $i<j$. 

For this, given any true parameter $\boldsymbol{\gamma}(i)$ we use an $m$-times forward pass-through the neural networks $F^1_{\phi}$ and $F^2_{\psi}$ to generate the approximate posterior of the neural network $\left(\hat{\boldsymbol{\gamma}}_j(i)\right)_{j=1}^m \sim \mathbb{Q}_{\phi}(\cdot\mid\boldsymbol{Z}^{[i]})$ and $\left(\hat{\bm\theta}_j(h_{\Delta})(i)\right)_{j=1}^m $. The optimal network parameters $\phi^*$ and $\psi^*$ are obtained by minimizing the empirical energy score in $\phi$ and $\psi$ across all simulation rounds $i\in \{1, \ldots, n\}$:
\begin{align}
\label{eq:energy_score_estimation}
    \textrm{S}_{\phi}(\bm \hat{\bm \gamma} (i), \boldsymbol{\gamma}(i)) & = \frac{1}{m} \sum_{j=1}^m \| \bm \hat{\bm \gamma}_j (i) - \bm \gamma(i)  \|_2 - \frac{1}{2 m (m-1)} \sum_{\substack{j,k = 1 \\ k \neq j}}^m \| \bm \hat{\bm \gamma}_j(i) - \bm \hat{\bm \gamma}_k(i) \|_2. \\
    \textrm{S}_{\psi}(\hat{\bm\theta}_j(h_{\Delta})(i), {\bm\theta}_j(h_{\Delta})(i)) & =  \frac{1}{m} \sum_{j=1}^m \| \hat{\bm\theta}_j(h_{\Delta})(i) - {\bm\theta}_j(h_{\Delta})(i)  \|_2 - \frac{1}{2 m (m-1)} \sum_{\substack{j,k = 1 \\ k \neq j}}^m \| \hat{\bm\theta}_j(h_{\Delta})(i) - \hat{\bm\theta}_k(h_{\Delta})(i) \|_2. \nonumber
\end{align}
The idea of using proper scoring rules to train (parameters of) generative neural networks is based on recent findings by \cite{Pacchiardi.2021} and \cite{Chen.2022} for multivariate probabilistic forecasting. Note that the criterion \eqref{eq:energy_score_estimation} is an unbiased estimator \citep{pacchiardi2022likelihoodfree} of the general energy score $\mathrm{ES}(\p, \boldsymbol{y}) = \e \big[\| \boldsymbol{Y} - \boldsymbol{y} \| \big] - \frac{1}{2} \e \big[\| \boldsymbol{Y} - \boldsymbol{Y}' \| \big]$, with $\boldsymbol{Y}, \boldsymbol{Y'}$ iid draws of $\p$, where a closed-form solution is usually not admissible. While in principle many choices of scoring rules are available, we focus on the energy score that admits the multivariate case and is most commonly applied in multivariate probabilistic forecasting. The energy score is strictly proper and has a unique minimum under mild regularity conditions; moreover, it is rotation and shift invariant \citep{szekely2013energy} and therefore tailored to the geometry of the considered model classes. For the case of full distributional learning, the energy score has also been shown to be robust to out-of distribution scenarios \citep{shen2024engression}. The energy score is attractive for its simplicity. It can be computed directly from samples, unlike the Kullback-Leibler divergence or the Wasserstein distance, which generally lack sample-based estimators.

\begin{figure}[htb]
\includegraphics[width = \textwidth]{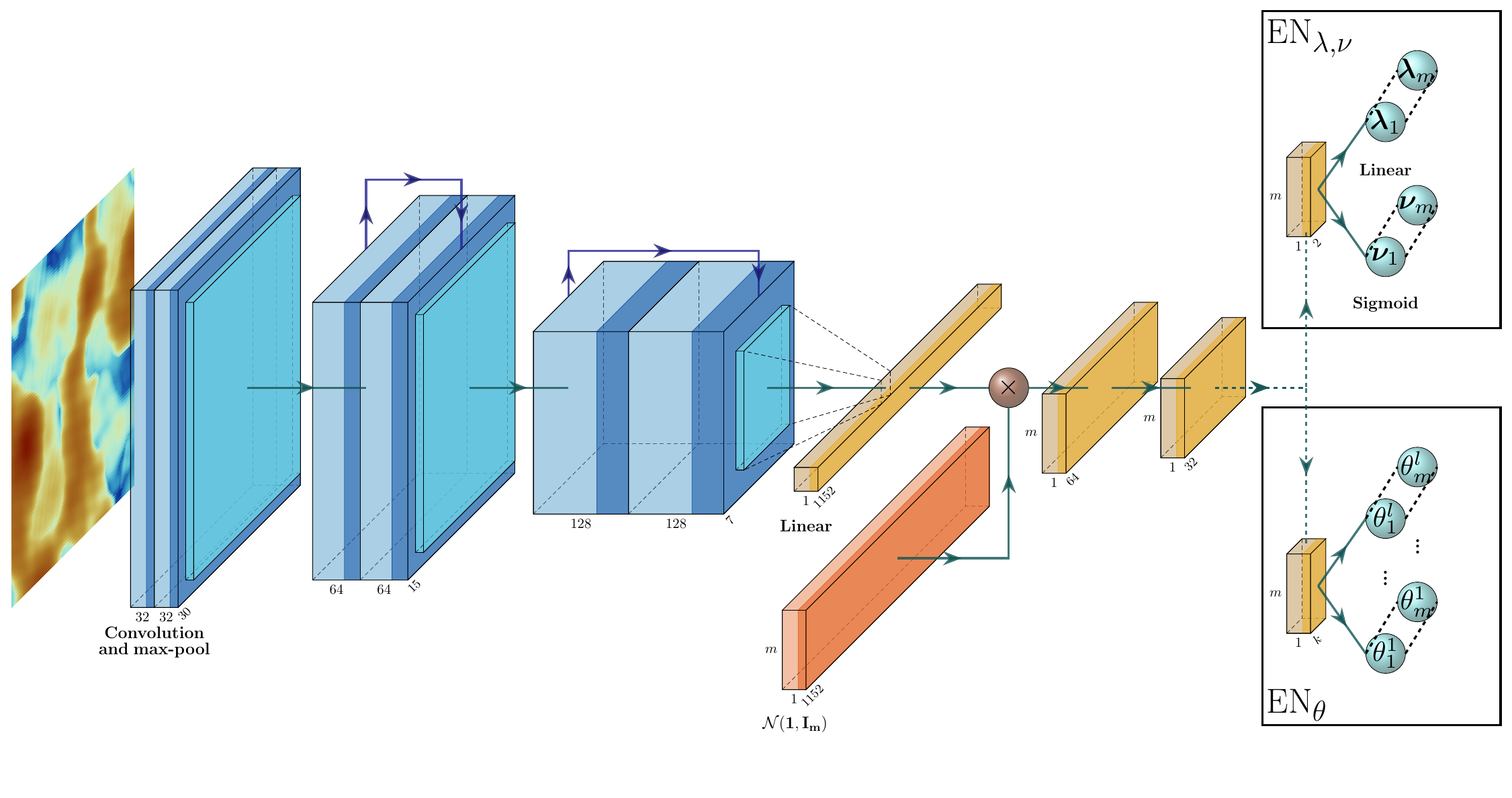}
\caption{The figure shows the proposed model architecture. The spatial field is fed through three blocks of convolutional and max-pooling layers. Across the blocks, the output size decreases, while the channel size increases. In the second and third block, residual connections are added, marked by the arrows on top. After the convolutional layers the network is flattened and fed through some final linear layers, where Gaussian noise is multiplied on top to create $m$ output samples. For parameter prediction, samples of $\gamma=(\lambda, \nu)'$ are created, while for the direct estimation of the pairwise extremal coefficient function, sample points of the function are predicted as $\theta^i_j := \hat{\theta}_j(h_i)$.}
\label{fig:model_architecture}
\end{figure}

For both network architectures of $F^1_\phi$ and $F^2_\psi$, we use convolutional neural networks (CNNs) that work well with inputs from time series of two-dimensional spatial data on a regular grid and have already been successfully applied to max-stable processes \citep{Lenzi.2023b, SainsburyDale.2022}. \autoref{fig:model_architecture} shows a visualization of the network. As the $\nu$ component  of $\gamma$ only takes values in $(0,2]$, we transform it to the unit interval and use a sigmoid activation function. For the $\lambda$ component of $\gamma$ we employ a log-transform and a linear activation function, similar to \cite{Lenzi.2023b}, while for the values of $\theta(h_1),\ldots,\theta(h_l)$ no transformation is required and we use a sigmoid activation function scaled to the range $(1,2)$. Note that for the $m$ samples of the posterior distribution of each target, the method samples from a latent space $\mathcal{N}(\boldsymbol{1}, \boldsymbol{I}_m)$ and multiplies the result to a linear layer (compare \autoref{fig:model_architecture}).
While different methods have been proposed to make training more efficient, such as using an informative prior \citep{Lenzi.2023b} or simulating new data during training \citep{SainsburyDale.2022}, we focus on techniques from image augmentation \citep{Perez.2017} that align with the symmetry of the underlying models in order to increase the training set $\mathbf{Z}$. We use image rotation of 180° and vertical and horizontal image flips with fixed probabilities that maintain the distances between spatial locations without distorting them and thus correspond to the stationary and isotropic Brown-Resnick and Schlather models. More details regarding the model architecture and implementation are available in \autoref{app:architecture}.

In the final prediction step, we use the pre-trained networks $F^1_{\phi^*}$ and $F^2_{\psi^*}$, that we denote as energy networks in the sequel,  on the available observed data set and obtain an empirical distribution of predictions $\hat{\bm\gamma}_1, \ldots, \hat{\bm\gamma}_m$ and $\hat{\bm \theta}_1(h_{\Delta}),\ldots, \hat{\bm \theta}_m(h_{\Delta})$. From each of these distributions, we can generate respective final point estimates $\hat{\bm\gamma}$ and $\hat{\bm \theta}(h_{\Delta})$ as pointwise sample averages. Correspondingly, we can also obtain pointwise prediction intervals by using pointwise empirical quantiles in order to provide respective measures of uncertainty. Note that for the final estimator $\hat{\bm \theta}(h_{\Delta})$ we suggest to use sorting either at the component level or for the aggregated estimator to ensure monotonicity with ascending spatial distance, i.e. $\hat{\theta}_r(h_i)\leq \hat{\theta}_r(h_j)$ for $i\leq j$ either at each component $r=1,\ldots,m$ or for the mean functional $\hat{\theta}$. More details can be found in \autoref{app:sorting}.

\subsection{Evaluation measures} \label{subsec:eva}
For assessing the adequacy of the final estimates, we propose to use measures for the mean and interval predictions of $\bm \gamma$ and $\theta(h)$, as well as for the respective predictive distributions. By providing these different metrics, we can assess the predictive performance of the point estimators while simultaneously analyzing the uncertainty in the predictions. Additionally, we assess the empirical coverage of the prediction intervals for different nominal levels.

For the point predictions of parameters $\bm\gamma$, we employ the typical mean squared error (MSE) as a metric, which we denote by $\text{MSE}_\gamma$ for the corresponding parameters $\gamma \in \{\lambda, \nu \}$. For the evaluation of the prediction intervals, we utilize the interval score \cite[IS][]{Gneiting.2007} which measures the fit of the predictive interval and the observation and which we denote as $\text{IS}_{\alpha, \gamma}$ for an interval of length $1-\alpha$ and parameter $\gamma$. Finally, to evaluate the predicted distribution $\mathbb{Q}_{\phi}(\cdot|\boldsymbol{Z})$ of the model with respect to the true parameters, we can use the already established energy score \eqref{eq:energy_score_estimation}. To assess the coverage of the posterior distribution estimates, we calculate the percentage of times the true parameter values $\lambda$ and $\nu$ fall within the estimated 95\%, 75\% and 50\% posterior intervals. For assessing the estimates of the dependence structure, we use pointwise analogues of the suggested metrics for $\theta(h;\bm \gamma)$. We calculate $\text{MSE}_\theta$ as the aggregate of the pointwise mean squared error of the pairwise extremal coefficient function on the grid $h_{\Delta}$. We use the grid-points $h_{\Delta}=(h_1, \ldots, h_l)$ to approximate an integral and therefore employ the trapezoidal rule in the aggregation. Similarly, we can generalize the interval score to evaluate pointwise confidence intervals for $\theta(h_{\Delta})$. For a fixed distance $h$, the interval score can be calculated by taking the empirical $\alpha$-quantile of the functions $\theta(h;\hat{\bm \gamma}_i), \ i=1,..,m$. The interval score over the entire function, is then given as the aggregate of interval scores over the grid $h_{\Delta}$, which we denote as integrated interval score $\text{IIS}_\alpha$ with respect to the level $\alpha$. The coverage of the extremal coefficient function is as well evaluated for each of the grid points $h_\Delta$. Finally, the energy score can be extended in a straightforward manner by plugging in the support points of the pairwise extremal coefficient function. 
%
%

\section{Simulation studies}
\label{sec:simulation}
In this section, we want to investigate the performance of our proposed approach (see \autoref{fig:method_workflow}) in finite samples. We study standard baseline scenarios and then look at different ways of model misspecification in order to highlight the robustness of the approach. Further details regarding the training details and hardware, as well as a runtime analysis of the methods are available in \autoref{app:architecture} and \autoref{app:runtime}, respectively.
Reproducible code is available at \url{https://github.com/cbuelt/spatio-temporal-extremes}.

\subsection{Setup}
\label{ssec:model}

\begin{figure}
    \centering
    \includegraphics[width = \linewidth]{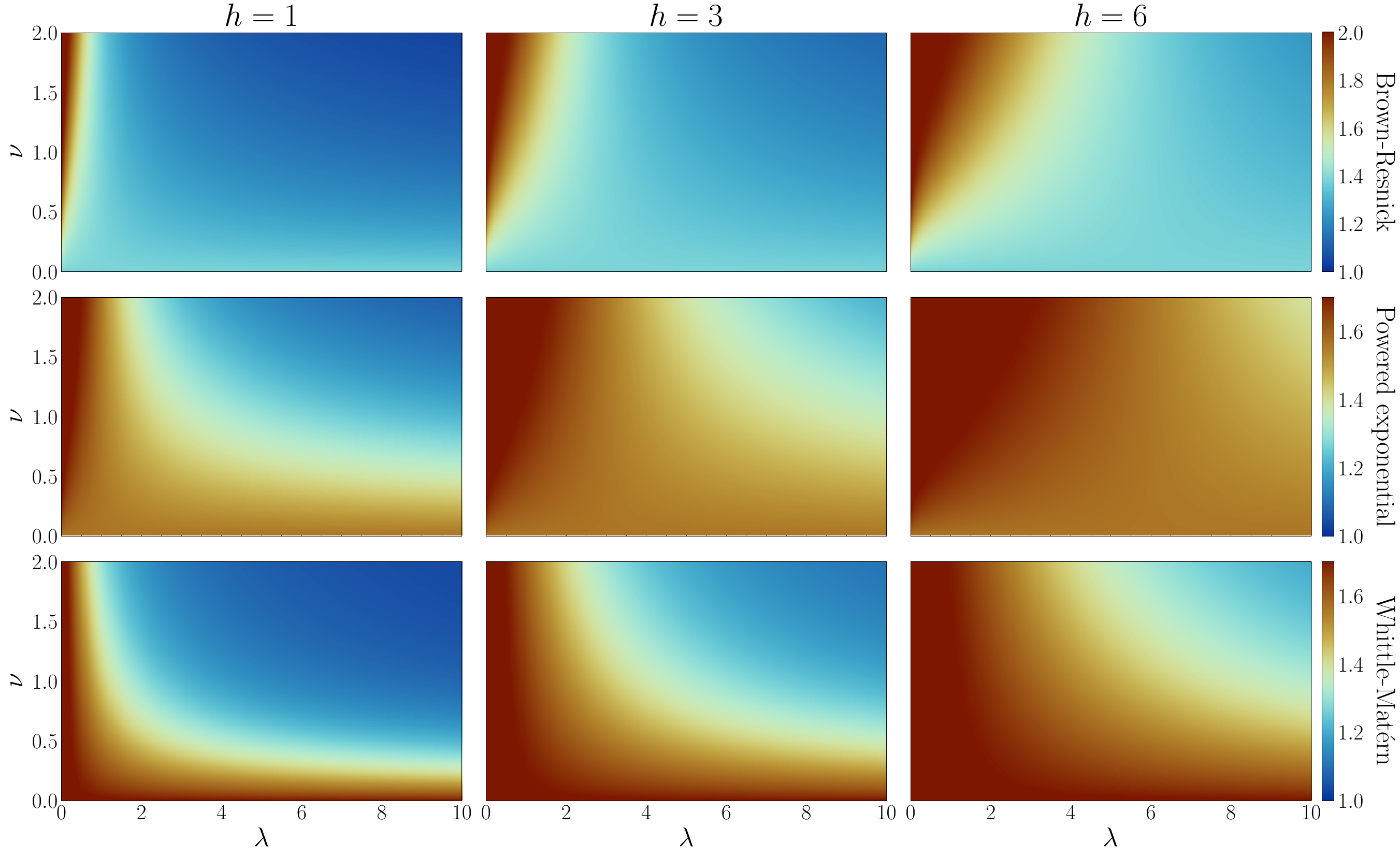}
    \caption[Visualization of the pairwise extremal coefficient function]{Visualization of the pairwise extremal coefficient function in dependence of $\lambda$ and $\nu$ for different models and distances $h$. In the top row, $h$ is set to $1$, in the middle $h=3$, and in the bottom row $h=6$.}
    \label{fig:visualization_ext_coef_func}
\end{figure}

We use the previously introduced Brown-Resnick and Schlather powered exponential model with $k=900$ spatial locations evenly distributed on the domain $\mathcal{D} = [0,30]^2$. In particular, for the baseline settings, we generate processes from both types of max-stable models via the parameters
\begin{equation}   
\label{eq:test_set}
(\lambda_i^{\mathrm{test}}, \nu_i^{\mathrm{test}}): \ \lambda_i^{\mathrm{test}} \sim \mathcal{U}(0.5,5), \ \nu_i^{\mathrm{test}} \sim \mathcal{U}(0.3,1.8), \quad i = 1, \ldots, 250,
\end{equation}
using the R-package \emph{SpatialExtremes} \citep{Ribatet.2022} as ground truth and use the same, correct type of max-stable model in the training of the network. In the following robustness scenarios, this will be relaxed. 
Note that the 250 baseline cases in the proposed setup cover a wide range of parameter combinations and different spatial dependencies (see \autoref{fig:visualization_ext_coef_func}). 

The training set of size $n=5000$ is generated from the same parameter space, with $20\%$ of the data used as validation data. For the direct estimation of $\boldsymbol{\theta}(h_{\Delta})$ we use an upper bound of h as $h \in (0, \sqrt{30^2+30^2}]$ and a grid spacing $\Delta = 0.1$. The neural network, with the proposed architecture as in \autoref{fig:model_architecture}, is trained by minimizing the energy score. A detailed description of the model architecture and training process is available in \autoref{app:architecture}. For the remainder of this article, we will refer to both of these networks as the energy networks EN$_{\lambda, \nu}$ and EN$_\theta$ for the parameter estimation and direct estimation, respectively. Note that both networks share the same hyperparameters and architectures, except for the final layer.

Through all settings, we implement several benchmark methods as comparison. First, we use the previously described pairwise likelihood (PL) method, following the setup of \cite{Lenzi.2023}, where the optimizer is run from 20 starting values, from which the 5 best estimates are again used as starting values, leading to the final estimate. The weight cutoff is chosen as $5$ and the estimator is fitted using the function \emph{fitmaxstab} of the R-package \emph{SpatialExtremes} \citep{Ribatet.2022}. Furthermore, we employ the approximate Bayesian computation (ABC) method. Following \cite{Erhardt.2012}, we use the proposed tripletwise extremal coefficient as a summary statistic, with a downsampled grid of size $5\times 5$ due to computational feasibility. We generate $50000$ independent simulations with $25$ processes each to compare against the observed data, where the cutoff is chosen such that the algorithm results in $m=500$ samples.
Further, we compare against two neural network baselines. First, we consider a ``regular'' implementation of the CNN with the same hyperparameters but minimizing the mean squared error, which is the setting in \cite{Lenzi.2023b}. In addition, we consider a predictive Gaussian approximation of the parameters, optimized with the log-score, which we refer to as $\mathrm{Log}_{\lambda, \nu}$. This allows for comparing our approach against a typically used parametric model. Further details of this model, as well as a comparison against a multivariate normal method trained with the energy score is available in \autoref{app:log_score}.

\subsection{Simulation results}
\begin{table}[ht]
\centering
\caption[Split results]{Table a) shows averages of probabilistic performance scores (see Section \ref{subsec:eva}) for the obtained empirical distribution of the posterior of $\gamma$ and $\theta$ across all 250 baseline scenarios of each of the two model types for the different estimation methods and evaluation metrics. Table b) depicts results for pointwise errors of the respective pointwise estimates of $\gamma$ and $\theta$. All metrics are negatively oriented, with the best model highlighted in bold and respective standard deviations are given in brackets. Table c) shows the empirical coverage of the estimated intervals for different levels, where the optimal value is the corresponding interval width.}
\label{tab:split_results}
\footnotesize

\caption*{\textbf{(a) Pointwise evaluation of average estimators}}
\begin{tabular}{|c|c|c|c|c|}
\hline
\textbf{Model} & \textbf{Estimator} & \textbf{MSE\textsubscript{$\lambda$}} & \textbf{MSE\textsubscript{$\nu$}} & \textbf{MSE\textsubscript{$\theta$}} \\
\hline
\multirow{6}{*}{\textbf{Brown-Resnick}} 
& EN$_{\lambda,\nu}$ & \textbf{0.38 (0.85)} & \textbf{0.01 (0.02)} & \textbf{0.15 (0.25)} \\ 
\cline{2-5}
& Log$_{\lambda,\nu}$  & 0.43 (0.88) & 0.02 (0.03) & 0.17 (0.29) \\
\cline{2-5}
& CNN & 0.43 (0.89) & 0.02 (0.04) & 0.19 (0.34) \\ 
\cline{2-5}
& PL & 3.19 (20.44) & 0.13 (0.22) & 0.71 (1.20) \\ 
\cline{2-5}
& ABC & 1.61 (1.79) & 0.26 (0.32) & 1.33 (1.67) \\ 
\cline{2-5}
& EN$_\theta$ & - & - & 0.16 (0.27) \\ 
\hline
\hline
\multirow{6}{*}{\textbf{Powexp}} 
& EN$_{\lambda,\nu}$ & 0.66 (1.24) & 0.05 (0.08) & 0.02 (0.03) \\ 
\cline{2-5}
& Log$_{\lambda,\nu}$  & \textbf{0.47 (1.07)} & \textbf{0.02 (0.04)} & \textbf{0.01 (0.02)} \\
\cline{2-5}
& CNN & \textbf{0.47 (0.96)} & 0.03 (0.05) & \textbf{0.01 (0.02)} \\ 
\cline{2-5}
& PL & 8.35 (6.86) & 0.78 (0.85) & 0.29 (0.17) \\ 
\cline{2-5}
& ABC & 1.60 (1.49) & 0.19 (0.17) & 0.05 (0.04) \\ 
\cline{2-5}
& EN$_\theta$ & - & - & \textbf{0.01 (0.02)} \\ 
\hline
\end{tabular}

\caption*{\textbf{(b) Probabilistic evaluation of the posterior distribution estimates}}
\begin{tabular}{|c|c|c|c|c|c|c|}
\hline
\textbf{Model} & \textbf{Estimator} & \textbf{IS\textsubscript{0.05,$\lambda$}} & \textbf{IS\textsubscript{0.05,$\nu$}} & \textbf{ES\textsubscript{$\lambda,\nu$}} & \textbf{IIS\textsubscript{0.05}} & \textbf{ES\textsubscript{$\theta$}} \\
\hline
\multirow{4}{*}{\textbf{Brown-Resnick}} 
& EN$_{\lambda,\nu}$ & 3.03 (6.43) & 0.55 (0.49) & \textbf{0.34 (0.32)} & \textbf{10.66 (14.19)} & 1.89 (0.78) \\ 
\cline{2-7}
& Log$_{\lambda,\nu}$  & \textbf{2.89 (5.41)} & \textbf{0.53 (0.77)} & 0.35 (0.32) & 12.03 (19.90) & 1.88 (0.84) \\
\cline{2-7}
& ABC & 4.59 (3.03) & 3.43 (6.53) & 0.85 (0.43) & 53.83 (93.44) & 1.92 (0.68) \\ 
\cline{2-7}
& EN$_\theta$ & - & - & - & 13.71 (23.00) & \textbf{0.74 (0.59)} \\ 

\hline
\hline
\multirow{4}{*}{\textbf{Powexp}} 
& EN$_{\lambda,\nu}$ & 3.45 (4.05) & 1.35 (2.21) & 0.46 (0.36) & 2.98 (3.67) & 0.47 (0.14) \\ 
\cline{2-7}
& Log$_{\lambda,\nu}$ & \textbf{3.23 (6.97)} & \textbf{0.61 (0.86)} & \textbf{0.36 (0.33)} & \textbf{2.31 (6.01)} & 0.45 (0.16) \\
\cline{2-7}
& ABC & 4.30 (0.33) & 1.47 (0.26) & 0.82 (0.32) & 4.22 (3.28) & 0.45 (0.16) \\ 
\cline{2-7}
& EN$_\theta$ & - & - & - & 3.58 (5.09) & \textbf{0.23 (0.14)} \\ 
\hline
\end{tabular}

\caption*{\textbf{(c) Coverage of the posterior distribution estimates}}
\begin{tabular}{|c|c|c|c|c|c|c|c|c|c|c|}
\hline
\multirow{2}{*}{\textbf{Model}} &\multirow{2}{*}{\textbf{Estimator}}  & \multicolumn{3}{c|}{$ 95\%$} & \multicolumn{3}{c|}{$75\%$} & \multicolumn{3}{c|}{$50\%$}\\
\cline{3-11}
 &  & $\lambda$ & $\nu$ & $\theta$ & $\lambda$ & $\nu$ & $\theta$ & $\lambda$ & $\nu$ & $\theta$\\
\hline
\multirow{4}{*}{\textbf{Brown-Resnick}} 
& EN$_{\lambda,\nu}$ & 0.81 & 0.91 & 0.84 & 0.57 & 0.67 & 0.63 & 0.42 & 0.44 & 0.40 \\
\cline{2-11}
& Log$_{\lambda,\nu}$ & 0.93 & \textbf{0.94} & \textbf{0.87} & 0.69 & \textbf{0.69} & \textbf{0.65} & 0.40 & \textbf{0.49} & 0.40 \\
\cline{2-11}
& ABC & \textbf{0.94} & 0.80 & 0.78 & \textbf{0.71} & 0.53 & 0.57 & \textbf{0.50} & 0.35 & 0.36 \\
\cline{2-11}
& EN$_\theta$ & - & - & 0.85 & - & - & 0.64 & - & - & \textbf{0.41} \\

\hline
\hline
\multirow{4}{*}{\textbf{Powexp}} 
& EN$_{\lambda,\nu}$ & 0.90 & 0.89 & 0.89 & 0.69 & 0.66 & 0.69 & 0.47 & 0.44 & \textbf{0.51} \\
\cline{2-11}
& Log$_{\lambda,\nu}$ & 0.92 & 0.91 & 0.88 & 0.68 & 0.68 & 0.67 & 0.43 & 0.44 & 0.46 \\
\cline{2-11}
& ABC& \textbf{0.96} & \textbf{0.94} & \textbf{0.97} & \textbf{0.76} & \textbf{0.73} & 0.83 & \textbf{0.51} & \textbf{0.51} & 0.56 \\
\cline{2-11}
& EN$_\theta$ & - & - & 0.91 & - & - & \textbf{0.78} & - & - & 0.59 \\
\hline
\end{tabular}

\end{table}

\begin{figure}[ht]
    \centering
    \begin{subfigure}[b]{\textwidth}
        \caption{Brown-Resnick}
    \includegraphics[width=\textwidth]{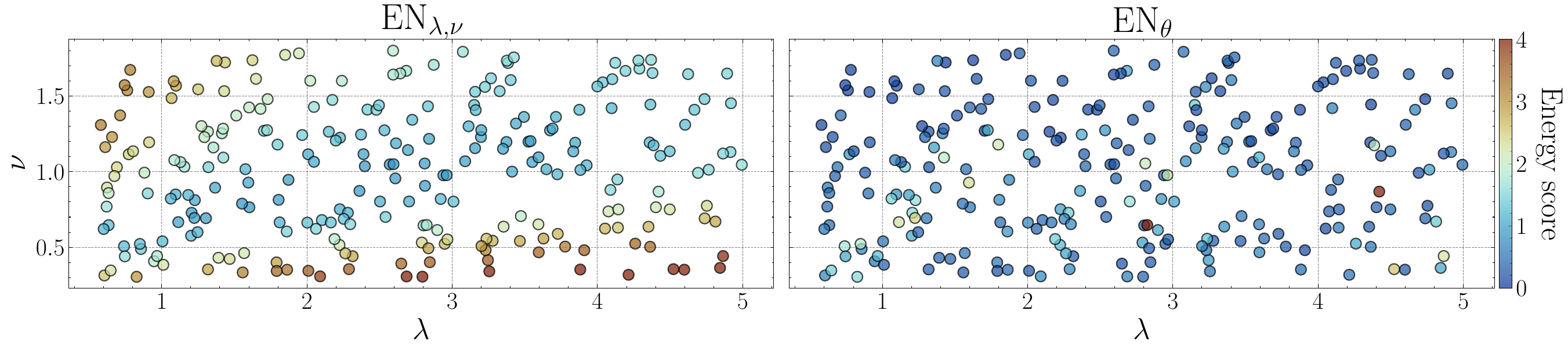}    \label{fig:energy_scores_normal_brown}
    \vspace{-0.7cm}
    \end{subfigure}
    \begin{subfigure}[b]{\textwidth}
        \caption{Powered exponential}
    \includegraphics[width=\textwidth]{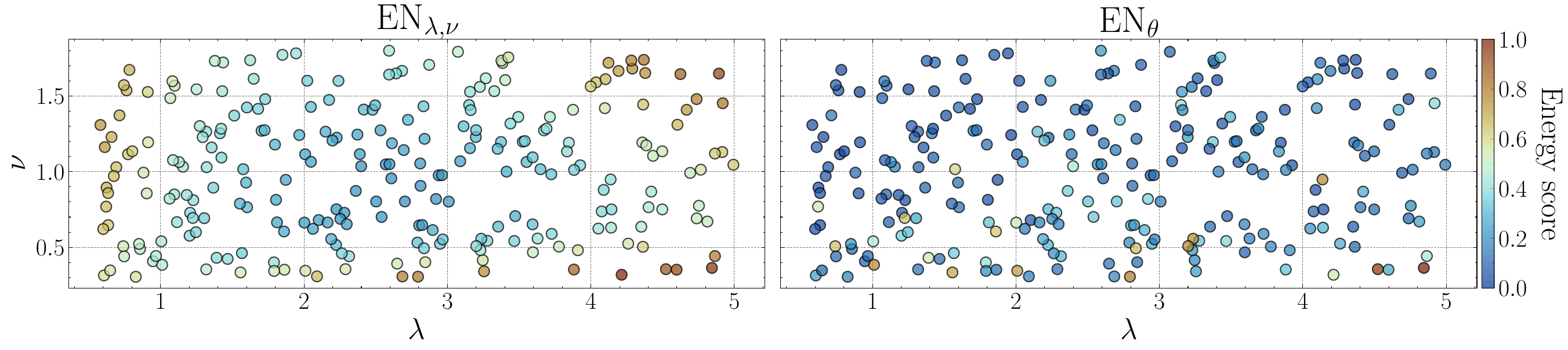}
    \label{fig:energy_scores_schlather}
        \vspace{-0.5cm}
    \end{subfigure}
    \caption{Visualization of the energy score across the parameters $(\lambda,\nu)$ from the test data for the Brown-Resnick and powered exponential model in the baseline scenario.}
    \label{fig:energy_scores_baseline}
\end{figure}

\begin{figure}[ht]
\includegraphics[width = \textwidth]{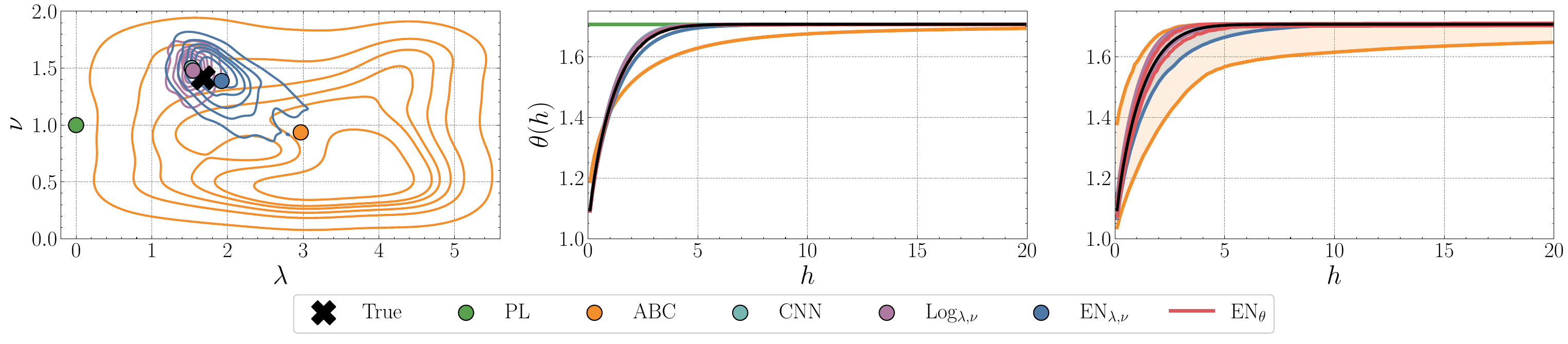}
\caption{The figure visualizes the different estimation methods for the max-stable models using a selected test sample of the powered exponential model with $(\lambda, \nu) = (1.68, 1.42)$. The left panel shows the different location estimates, along with contour plots for the predicted probability density of the ABC, EN$_{\lambda,\nu}$ and Log$_{\lambda,\nu}$  methods. The middle panel displays the estimated pairwise extremal coefficient functions, while the right panel shows the estimated pointwise confidence intervals ($95\%$) for the pairwise extremal coefficient function.}
\label{fig:results_single}
\end{figure}

The aggregate results of the baseline scenarios are presented in \autoref{tab:split_results}. Overall, the energy network exhibits the best performance for both types of max-stable models across all metrics regarding the pairwise extremal coefficient functions. For the Brown–Resnick model, the Energy Network produces better point estimates, while the log-score yields better interval estimates, as indicated by the interval score. This is not surprising, as under the true model, the normality assumption can be justified and the log-score leads to the minimum variance estimator in that case.  Nevertheless, the results in \autoref{app:log_score} show that these advantages vanish when combining a normality assumption with energy score training.

For the powered exponential model, the mean estimates of the standard CNN, the Log$_{\lambda,\nu}$, and the Energy Network are very similar, with a slight advantage for Log$_{\lambda,\nu}$. The coverage results suggest that, especially for the 95\% intervals, the Energy Network tends to underestimate uncertainty in the baseline scenario. Although the ABC method attains good coverage, its estimates exhibit high variability and its point estimation accuracy is poor, making the Energy Network the more attractive method overall.

The results also highlight that training separate networks for separate target parameters $\gamma$ and $\theta$ is crucial, in particular for distributional estimates. 
\autoref{fig:energy_scores_baseline} shows the energy score in dependence of the underlying parameters $(\nu, \theta)$. For both models, the EN$_\theta$ does not exhibit elevated energy scores for any particular values of $\lambda$ and $\nu$, while for EN$_{\lambda,\nu}$ the scores are higher close to the boundary of the test parameter range, as the estimates of EN$_{\lambda,\nu}$ are highly dependent on the parameters, as opposed to EN$_\theta$. A visualization of a specific randomly selected setting among the 250 baselines of the Brown-Resnick model is shown in \autoref{fig:results_single}, highlighting that the predicted parameter distribution of the energy network has a significantly lower spread, as compared to the ABC method. The same holds for the functional prediction and the corresponding prediction intervals, where $\mathrm{EN}_{\theta}$ also shows similar performance. Additional visualizations can be found in \autoref{app:additional_visualizations}.

\subsection{Robustness analysis}
In addition to the previous analysis, we want to further understand the robustness of the proposed estimation method. For that purpose, we provide the following three different simulation scenarios with different types of misspecification:
\begin{enumerate}
    \item \textbf{Misspecified parameter range:} In the first scenario, we analyze how the energy network performs under a misspecification of the parameter space in the setting of a Brown-Resnick model. For that purpose, the set of parameters in the training data is chosen disjoint from the test data, forcing the models to infer out-of-distribution predictions. The training range is chosen as $\lambda^{\mathrm{train}} \in [0.5,5], \ \nu^{\mathrm{train}} \in [0.3,1.8]$, while the test set covers $\lambda^{\mathrm{test}} \in (0,0.5) \cup (5,10], \ \nu^{\mathrm{test}} \in (0,0.3) \cup (1.8,2]$.
    \item \textbf{Misspecified correlation function:} The second scenario analyzes the performance under a misspecified correlation function, in this case by using a misspecified Schlather process. The test set is generated with the same parameters as before for a Schlather model with a Whittle-Matérn kernel, while the training set is generated using a powered exponential kernel. 
    \item \textbf{Misspecified model:} Finally, we investigate whether the methods are robust against general model misspecification. For this purpose, the methods are trained on a Brown-Resnick model, while the true processes stem from a Smith model, which is a special case of the former, as mentioned earlier. More specifically, consider a Smith model with diagonal covariance matrix $\Sigma = \textrm{diag}(\sigma)$, which corresponds to a Brown-Resnick process with $\nu = 2$ and $\lambda = \sqrt{2\sigma}$. An appropriate estimator should therefore always predict the smoothness parameter as $\nu = 2$. Again, a test set of size $n=250$ is simulated based on the Brown-Resnick model with $\lambda \sim \mathcal{U}(0.5,5), \ \nu\sim \mathcal{U}(0,2)$, while the true data is simulated from a Smith model with $\sigma \sim \mathcal{U}(0.5,5)$. 
\end{enumerate}
Evaluating these scenarios gives insights into how the energy network is able to extrapolate across the parameter range and illustrates its robustness in practice when the true model is generally unknown. These properties then help to understand estimation performance in real-world data settings. The numerical results for the different scenarios of misspecification are shown in \autoref{tab:robustness_split_results}.

Overall, as expected, scores and average errors are larger in the misspecified than in the baseline scenario. The domain misspecification is the hardest robustness check for all methods, according to the increases in scores and errors. 

\begin{figure}[ht]
    \centering
    \begin{subfigure}[b]{\textwidth}
        \caption{Misspecified parameter range}
    \includegraphics[width=\textwidth, keepaspectratio]{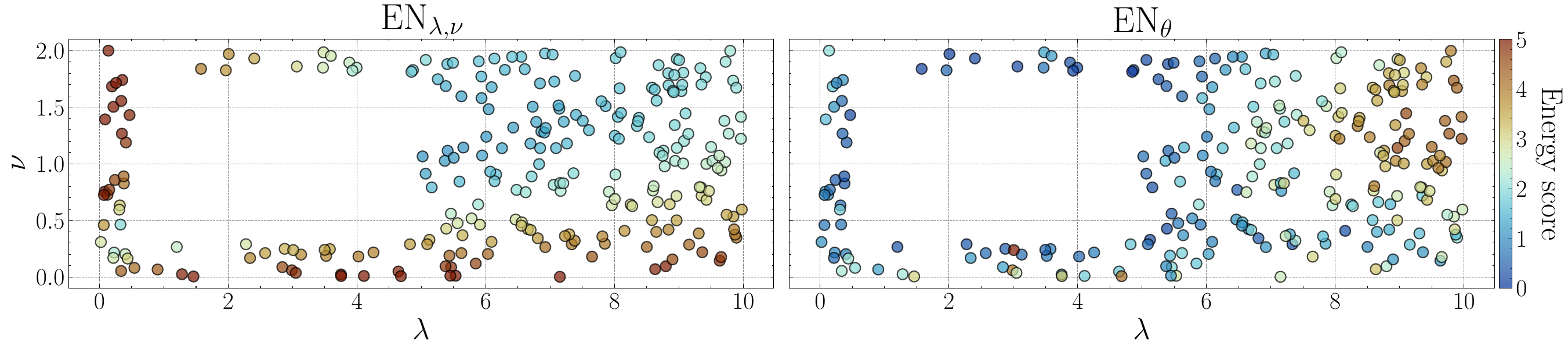}
    \label{fig:energy_scores_robustness_scenario_1}
    \vspace{-0.7cm}
    \end{subfigure}
    \begin{subfigure}[b]{\textwidth}
        \caption{Misspecified correlation function}
    \includegraphics[width=\textwidth, keepaspectratio]{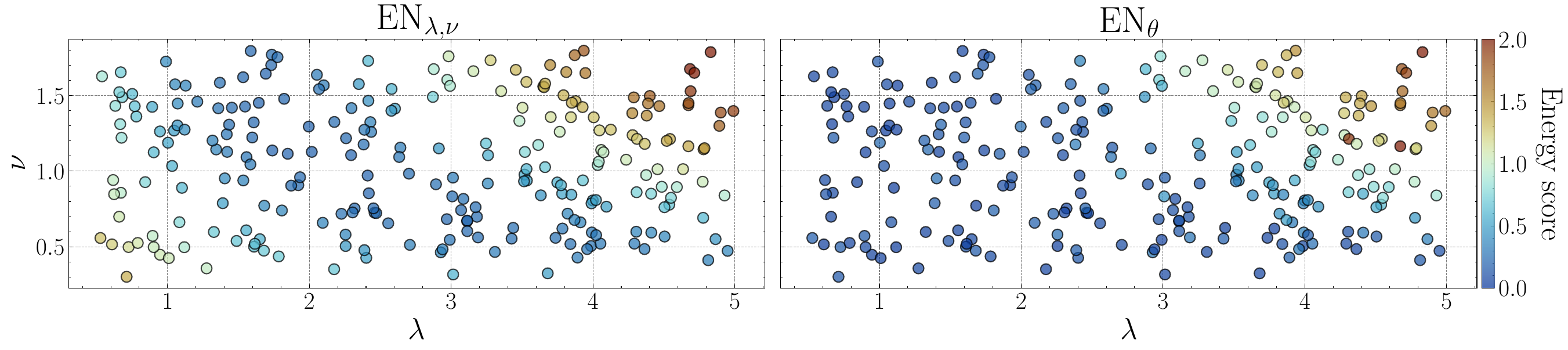}
        \label{fig:energy_scores_robustness_scenario_2}
        \vspace{-0.7cm}
    \end{subfigure}
        \begin{subfigure}[b]{\textwidth}
        \caption{Misspecified model}
    \includegraphics[width=\textwidth, keepaspectratio]{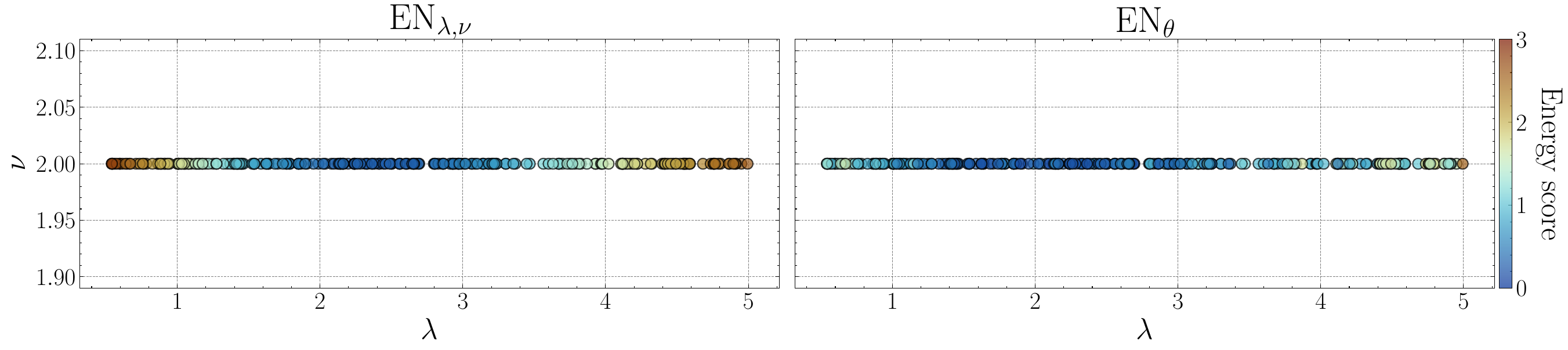}
        \label{fig:energy_scores_robustness_scenario_3}
        \vspace{-0.5cm}
    \end{subfigure}
    \caption{The figure visualizes the energy score across the parameters $(\lambda,\nu)$ from the test data for the Brown-Resnick and powered exponential model for three scenarios of misspecification.}
    \label{fig:energy_scores_robustness}
\end{figure}

\begin{figure}[ht]
\includegraphics[width = \textwidth]{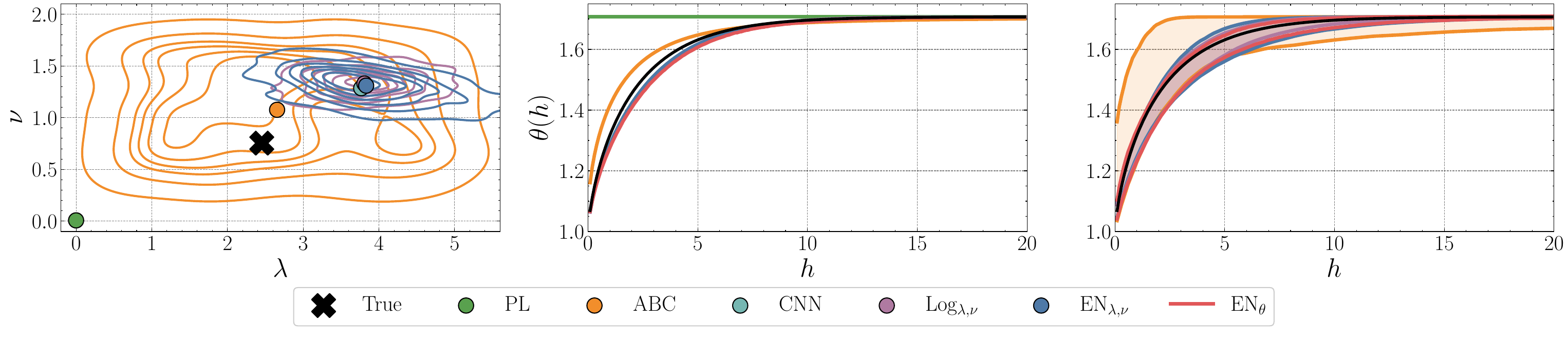}
\caption{The figure visualizes the different estimation methods for a misspecified correlation function, where the network is trained on a powered exponential model but predicts a Whittle-Matérn model with $(\lambda, \nu) = (2.45, 0.75)$. Further specifications are as in \autoref{fig:results_single}}
\label{fig:results_single_robust}
\end{figure}

\begin{table}[h]
\centering
\caption[Results]{The table shows the metrics for the evaluation of the extremal spatial dependence for the combined model training for different underlying truths based on 1666 and 5000 datapoints each. All metrics are negatively oriented, with the best model highlighted in bold and standard deviation given in brackets.}
\label{tab:aggregated_data_results_small}
\begin{tabular}{|c|c|c|c|c|c|}
\hline
\textbf{Model} & $\bm n$ & \textbf{Estimator} & \textbf{MSE\textsubscript{$\theta$}} & \textbf{IIS\textsubscript{0.05}} & \textbf{ES\textsubscript{$\theta$}} \\
\hline

\multirow{6}{*}{\textbf{Brown-Resnick}} 
& \multirow{3}{*}{\textbf{1666}} 
& CNN & 0.33 (0.6) & - & - \\
\cline{3-6}
& & EN$_{\lambda,\nu}$ & 0.37 (0.63) & \textbf{15.94 (19.62)} & 1.91 (0.96) \\
\cline{3-6}
& & EN$_\theta$ & \textbf{0.29 (0.45)} & 20.96 (33.60) & \textbf{1.00 (0.79)} \\
\cline{2-6}
& \multirow{3}{*}{\textbf{5000}} 
& CNN & 0.20 (0.35) & - & - \\
\cline{3-6}
& & EN$_{\lambda,\nu}$ & 0.23 (0.40) & \textbf{13.94 (19.45)} & 1.89 (0.91) \\
\cline{3-6}
& & EN$_\theta$ & \textbf{0.17 (0.30)} & 15.39 (32.37) & \textbf{0.74 (0.66)} \\
\hline

\multirow{6}{*}{\textbf{Powexp}} 
& \multirow{3}{*}{\textbf{1666}} 
& CNN & \textbf{0.02 (0.02)} & - & - \\
\cline{3-6}
& & EN$_{\lambda,\nu}$ & \textbf{0.02 (0.03)} & \textbf{3.70 (8.13)} & 0.47 (0.20) \\
\cline{3-6}
& & EN$_\theta$ & 0.03 (0.04) & 5.80 (8.16) & \textbf{0.35 (0.22)} \\
\cline{2-6}
& \multirow{3}{*}{\textbf{5000}} 
& CNN & 0.02 (0.02) & - & - \\
\cline{3-6}
& & EN$_{\lambda,\nu}$ & 0.02 (0.02) & \textbf{2.57 (2.51)} & 0.46 (0.18) \\
\cline{3-6}
& & EN$_\theta$ & \textbf{0.01 (0.02)} & 2.76 (3.63) & \textbf{0.21 (0.16)} \\
\hline
\end{tabular}

\end{table}

\begin{table}[ht]
\centering
\caption[Robustness split results]{
Table a) shows averages of probabilistic performance scores (see Section \ref{subsec:eva}) for the obtained empirical distribution of the posterior across the three robustness settings for the different estimation methods and evaluation metrics. Table b) depicts results for pointwise errors of the respective pointwise estimates. All metrics are negatively oriented, with the best model highlighted in bold and respective standard deviations are given in brackets. Table c) shows the empirical coverage of the estimated intervals for different levels, where the optimal value is the corresponding interval width.}
\label{tab:robustness_split_results}
\footnotesize

\caption*{\textbf{(a) Pointwise evaluation of average estimators (Robustness Scenarios)}}
\begin{tabular}{|c|c|c|c|c|}
\hline
\textbf{Scenario} & \textbf{Estimator} & \textbf{MSE\textsubscript{$\lambda$}} & \textbf{MSE\textsubscript{$\nu$}} & \textbf{MSE\textsubscript{$\theta$}} \\
\hline

\multirow{6}{*}{\makecell{\textbf{Misspecified} \\ \textbf{parameter range}}} 
& EN$_{\lambda,\nu}$ & \textbf{9.47 (10.15)} & \textbf{0.03 (0.06)} & \textbf{0.55 (0.52)} \\ 
\cline{2-5}
& Log$_{\lambda,\nu}$ & 9.77 (10.35) & 0.04 (0.09) & 0.72 (0.67) \\
\cline{2-5}
& CNN & 9.68 (10.65) & 0.06 (0.10) & 0.78 (0.83) \\ 
\cline{2-5}
& PL & 13.07 (29.13) & 0.23 (0.55) & 0.91 (1.46) \\ 
\cline{2-5}
& ABC & 18.01 (15.15) & 0.52 (0.58) & 2.65 (2.48) \\ 
\cline{2-5}
& EN$_\theta$ & - & - & 0.76 (0.65) \\ 
\hline
\hline

\multirow{6}{*}{\makecell{\textbf{Misspecified} \\ \textbf{correlation function}}} 
& EN$_{\lambda,\nu}$ & 1.76 (1.87) & \textbf{0.16 (0.16)} & 0.04 (0.06) \\ 
\cline{2-5}
& Log$_{\lambda,\nu}$ & \textbf{1.32 (1.53)} & 0.19 (0.15) & 0.04 (0.07) \\
\cline{2-5}
& CNN & 2.37 (2.82) & 0.17 (0.17) & \textbf{0.03 (0.04)} \\ 
\cline{2-5}
& PL & 9.61 (7.28) & 0.94 (0.89) & 0.67 (0.42) \\ 
\cline{2-5}
& ABC & 1.70 (1.47) & 0.17 (0.16) & 0.17 (0.19) \\ 
\cline{2-5}
& EN$_\theta$ & - & - & 0.05 (0.09) \\ 
\hline
\hline

\multirow{6}{*}{\makecell{\textbf{Misspecified} \\ \textbf{model}}} 
& EN$_{\lambda,\nu}$ & 0.11 (0.21) & 0.02 (0.02) & 0.04 (0.04) \\ 
\cline{2-5}
&Log$_{\lambda,\nu}$ & \textbf{0.05 (0.07)} & \textbf{0.01 (0.02)} & \textbf{0.02 (0.02)} \\
\cline{2-5}
& CNN & 0.15 (0.20) & 0.05 (0.05) & 0.08 (0.07) \\ 
\cline{2-5}
& PL & 0.23 (1.12) & 0.16 (0.39) & 0.25 (0.84) \\ 
\cline{2-5}
& ABC & 0.89 (1.11) & 0.81 (0.68) & 1.83 (2.31) \\ 
\cline{2-5}
& EN$_\theta$ & - & - & 0.03 (0.03) \\ 
\hline
\end{tabular}

\caption*{\textbf{(b) Probabilistic evaluation of the posterior distribution estimates (Robustness Scenarios)}}
\begin{tabular}{|c|c|c|c|c|c|c|}
\hline
\textbf{Scenario} & \textbf{Estimator} & \textbf{IS\textsubscript{0.05,$\lambda$}} & \textbf{IS\textsubscript{0.05,$\nu$}} & \textbf{ES} & \textbf{IIS\textsubscript{0.05}} & \textbf{ES\textsubscript{$\theta$}} \\
\hline

\multirow{4}{*}{\makecell{\textbf{Misspecified} \\ \textbf{parameter range}}}
& EN$_{\lambda,\nu}$ & \textbf{59.76 (56.30)} & \textbf{1.22 (1.00)} & \textbf{2.24 (1.63)} & \textbf{54.54 (54.08)} & 2.75 (1.17) \\ 
\cline{2-7}
& Log$_{\lambda,\nu}$ & 63.61 (59.62) & 2.52 (3.47) & 2.34 (1.65) & 99.67 (90.11) & 2.89 (1.25) \\
\cline{2-7}
& ABC & 87.49 (68.32) & 9.12 (12.24) & 3.25 (1.75) & 111.21 (100.04) & 2.88 (0.99) \\ 
\cline{2-7}
& EN$_\theta$ & - & - & - & 92.36 (78.12) & \textbf{2.03 (1.16)} \\ 
\hline
\hline

\multirow{4}{*}{\makecell{\textbf{Misspecified} \\ \textbf{correlation function}}} 
& EN$_{\lambda,\nu}$ & 12.38 (14.42) & 4.68 (5.57) & 0.92 (0.54) & \textbf{7.21 (12.46)} & 0.66 (0.42) \\ 
\cline{2-7}
& Log$_{\lambda,\nu}$ & 18.87 (20.86) & 10.40 (6.96) & 0.93 (0.51) & 14.01 (21.08) & 0.68 (0.45) \\
\cline{2-7}
& ABC & \textbf{4.31 (0.41)} & \textbf{1.46 (0.29)} & \textbf{0.83 (0.32)} & 13.80 (16.79) & 0.84 (0.58) \\ 
\cline{2-7}
& EN$_\theta$ & - & - & - & 13.12 (18.78) & \textbf{0.45 (0.46)} \\ 
\hline
\hline

\multirow{4}{*}{\makecell{\textbf{Misspecified} \\ \textbf{model}}} 
& EN$_{\lambda,\nu}$ & 1.43 (1.34) & 2.18 (1.20) & 0.20 (0.13) & 3.36 (3.56) & 0.62 (0.34) \\ 
\cline{2-7}
& Log$_{\lambda,\nu}$ & \textbf{1.13 (1.39)} & \textbf{0.32 (0.91)} & \textbf{0.15 (0.10)} & \textbf{2.89 (3.43)} & 0.60 (0.38) \\
\cline{2-7}
& ABC & 4.22 (0.89) & 9.28 (14.58) & 0.89 (0.40) & 79.17 (146.55) & 2.11 (0.40) \\ 
\cline{2-7}
& EN$_\theta$ & - & - & - & 3.87 (2.98) & \textbf{0.35 (0.14)} \\ 
\hline

\end{tabular}

\caption*{\textbf{(c) Coverage of the posterior distribution estimates}}
\begin{tabular}{|c|c|c|c|c|c|c|c|c|c|c|}
\hline
\multirow{2}{*}{\textbf{Model}} &\multirow{2}{*}{\textbf{Estimator}}  & \multicolumn{3}{c|}{$95\%$} & \multicolumn{3}{c|}{$75\%$} & \multicolumn{3}{c|}{$50\%$}\\
\cline{3-11}
 &  & $\lambda$ & $\nu$ & $\theta$ & $\lambda$ & $\nu$ & $\theta$ & $\lambda$ & $\nu$ & $\theta$\\
\hline
\multirow{4}{*}{\makecell{\textbf{Misspecified} \\ \textbf{parameter range}}} 
& EN$_{\lambda,\nu}$ & 0.18 & \textbf{0.56} & 0.27 & 0.10 & \textbf{0.38} & 0.17 & 0.06 & \textbf{0.21} & 0.11 \\
\cline{2-11}
& Log$_{\lambda,\nu}$ & \textbf{0.23} & 0.41 & 0.25 & 0.09 & 0.24 & 0.12 & \textbf{0.07} & 0.10 & 0.08 \\
\cline{2-11}
& ABC & 0.14 & 0.46 & \textbf{0.42} & \textbf{0.11} & 0.27 & \textbf{0.24} & \textbf{0.07} & 0.19 & \textbf{0.14} \\
\cline{2-11}
& EN$_\theta$ & - & - & 0.30 & - & - & 0.17 & - & - & 0.10 \\

\hline
\hline
\multirow{4}{*}{\makecell{\textbf{Misspecified} \\ \textbf{correlation function}}} 
& EN$_{\lambda,\nu}$ & 0.54 & 0.28 & 0.47 & 0.38 & 0.13 & 0.29 & 0.22 & 0.07 & 0.19 \\
\cline{2-11}
& Log$_{\lambda,\nu}$ & 0.40 & 0.12 & 0.42 & 0.25 & 0.08 & 0.32 & 0.14 & 0.04 & 0.21 \\
\cline{2-11}
& ABC & \textbf{0.97} & \textbf{0.96} & \textbf{0.91} & \textbf{0.74} & \textbf{0.82} & \textbf{0.85} & \textbf{0.48} & \textbf{0.54} & \textbf{0.64} \\
\cline{2-11}
& EN$_\theta$ & - & - & 0.69 & - & - & 0.49 & - & - & 0.33 \\

\hline
\hline
\multirow{4}{*}{\makecell{\textbf{Misspecified} \\ \textbf{model}}} 
& EN$_{\lambda,\nu}$ & \textbf{0.95} & 0.00 & \textbf{0.85} & \textbf{0.79} & 0.00 & \textbf{0.58} & \textbf{0.53} & 0.00 & \textbf{0.40} \\
\cline{2-11}
& Log$_{\lambda,\nu}$ & 0.76 & \textbf{0.87} & 0.63 & 0.44 & \textbf{0.07} & 0.35 & 0.27 & 0.00 & 0.20 \\
\cline{2-11}
& ABC & 0.99 & 0.00 & 0.76 & 0.90 & 0.00 & 0.26 & 0.67 & 0.00 & 0.07 \\
\cline{2-11}
& EN$_\theta$ & - & - & 0.48 & - & - & 0.35 & - & - & 0.24 \\
\hline
\end{tabular}

\end{table}

\paragraph{Misspecified parameter range}
In this case, the findings in \autoref{tab:robustness_split_results} reveal that on average the $\mathrm{EN}_{\lambda, \nu}$ yields the lowest scores and errors for almost all tailored pointwise and distributional metrics. Moreover, $\mathrm{EN}_{\theta}$ is clearly superior in the functional energy score while pointwise in the MSE, the $\mathrm{EN}_{\lambda, \nu}$ is even better. The coverage is overall for all methods way too low, showing the importance of specifying a wide parameter range in application scenarios.
\autoref{fig:energy_scores_robustness_scenario_1} provides an intuition of the results by displaying the distribution of the energy score with respect to the parameters $(\lambda, \nu)$ of the underlying process. The energy score is higher for the test parameters further away from the training parameters. For the EN$_\theta$, parameters in the upper right corner correspond to higher energy scores. A possible explanation can be obtained by considering the values of the pairwise extremal coefficient function in \autoref{fig:visualization_ext_coef_func}, where the first column shows that there exists nearly a full dependency, which is not the case for lower values of $\lambda$ and $\nu$. For the EN$_{\lambda,\nu}$, the energy scores look similar to those in \autoref{fig:energy_scores_normal_brown}, although the values are slightly higher. \autoref{fig:robust_results_multiple} in \autoref{app:additional_visualizations} shows visualizations of  selected test samples. While individual point estimates for $\gamma$ can be quite diluted (but still much less than for the considered benchmarks), their distributions generally perform reasonably well in that they contain the true as opposed to the considered benchmark techniques. For $\theta$ also the point estimates of the proposed method show a good performance outperforming the benchmarks. For the distributional estimates in this case, this superior performance is even more pronounced. It appears that $\mathrm{EN}_{\lambda, \nu}$ automatically retracts to the true parameter, extrapolating to values outside of the training range of the model. 

\paragraph{Misspecified correlation function}
In this case, \autoref{tab:robustness_split_results} shows that while $\mathrm{EN}_{\lambda, \nu}$ leads to high errors in estimating the parameters in $\gamma$, it exhibits a good performance regarding the functional metrics for $\theta$. Generally, the approach yields high errors in all parameter $\gamma$ related metrics (including the energy score) and a very low error in all metrics with respect to $\theta$. The direct estimation does not lead to noticeable improvements, except in the functional energy score. Only the ABC method attains coverage levels close to the nominal ones, but it performs very poorly in point estimation. Among the methods that still deliver plausible point estimates, $\mathrm{EN}_{\theta}$ achieves the highest coverage. Overall, coverage is much better than in the misspecified parameter setting, but it remains substantially below that of the baseline scenario. Similar to the previous scenario, \autoref{fig:energy_scores_robustness_scenario_2} showcases that for the EN$_\theta$ the energy score is high whenever $\lambda$ and $\nu$ take on large values. \autoref{fig:visualization_ext_coef_func} shows that in this area in particular, the values for the powered exponential kernel and the Whittle-Matérn kernel differ significantly. The same effect can be seen for the EN$_{\lambda,\nu}$, which highlights that, depending on the given data and true parameter, one of the energy networks might be preferable over the other. While the energy scores for the EN$_\theta$ in the baseline scenario do not depend on parameter values, the EN$_{\lambda,\nu}$ performs worse on the margin of the trained parameter range. Visualizations for selected cases can be found in \autoref{app:additional_visualizations}. The $\mathrm{EN}_{\lambda, \nu}$ extrapolates to some previously unknown parameter range, leading to a poor parameter estimation of $\gamma$ but a good representation of the spatial dependence, as well as corresponding confidence intervals.

%
%
\paragraph{Misspecified model}
In this case, EN$_{\lambda,\nu}$ and Log$_{\lambda,\nu}$ have the lowest error for all parameter $\gamma$ related metrics, according to \autoref{tab:robustness_split_results}. Especially the error of $\nu$ is significantly lower as compared to the other methods, indicating that the EN$_{\lambda,\nu}$ is able to correctly identify the fixed parameter $\nu = 2$. The EN$_{\theta}$ leads to significant improvements regarding the metrics involving the pairwise extremal coefficient function. The coverage results show that EN$_{\lambda,\nu}$ performs well for the parameter $\lambda$ and for the extremal coefficient function. For $\nu$, however, the coverage is $0$, because the network constrains its predictions to the interval $(0,2)$ and therefore cannot include the boundary value $\nu=2$ in its prediction intervals. Nevertheless, the MSE indicates that the method still estimates this parameter accurately despite this limitation. \autoref{fig:energy_scores_robustness_scenario_3} shows that for the EN$_{\lambda, \nu}$ the energy scores are higher closer to the boundary of the parameter $\lambda$, while for the EN$_\theta$ the energy scores are higher only for very large $\lambda$. A visualization of selected test samples can be found in \autoref{fig:robust_results_multiple_smith} in \autoref{app:additional_visualizations}. It is evident that the predictive distribution of the EN$_{\lambda,\nu}$ is very concentrated, indicating that the model is confident that the true parameter lies in that range. This is also reflected in the narrow predictive intervals for the pairwise extremal coefficient function.

Motivated by the previous analysis of robustness against model misspecification, we also train a combined model on different types of max-stable processes simultaneously. For this purpose, training data is generated from the Brown-Resnick and the Schlather model with both, a powered exponential and Whittle-Matérn kernel. Combining simulated data from these models covers many different data scenarios; therefore, the network predictions should generalize across the different models. We consider two different training sets, containing 1666 (5000 split across three models) and 5000 datapoints for each model, respectively. The first case corresponds to the total data size of the previous scenarios, while the second case also captures the effect of an increased amount of data. The trained models are evaluated on the truth data from the baseline setting covering the Brown-Resnick model and from the Schlather model with powered exponential kernel. The results in \autoref{tab:aggregated_data_results_small} show that the metrics for both the EN$_{\lambda,\nu}$ and the EN$_\theta$ are improved when training on different models and an enlarged dataset. This is advantageous when dealing with real data, where the true model is unknown and cannot be used to determine the pairwise extremal coefficient function, as required for EN$_{\lambda,\nu}$. Altogether, these results suggest that the EN$_\theta$ might benefit from training on different max-stable models, as it can then predict the pairwise extremal coefficient function of a given dataset regardless of its true underlying model. 

In summary, by analyzing several types of misspecification, we cover different cases that are of relevance in a real-data scenario and show that the energy networks still produce robust and reliable results outperforming existing benchmarks. A closer look at the (functional) energy scores for single observations gives further insight into the cases where the estimation seems to perform better or worse. 

%
\FloatBarrier
\section{Modeling precipitation extremes}
\label{sec:application}
In this section, we apply our methodology to study extreme precipitation and the associated flooding events in Germany, focusing on two of the most severe incidents of the last decades: the July 2021 Ahr valley flood in western Germany and the August 2002 Elbe flood in eastern Germany. Since flooding events are driven by spatially concentrated precipitation extremes, accurately capturing the underlying spatial dependence structure is essential for risk assessment, including the derivation of return periods and durations \citep{https://doi.org/10.1029/2023WR034718,BENNETT20181123,https://doi.org/10.1002/2017WR022231}. In particular, as described in Section~\ref{sec:data}, we aggregate yearly summer precipitation maxima across all days at each grid point, a standard approach for obtaining max-stable processes \citep{SchlatherM.TawnA..2003}, which we use as an underlying spatial dependence model.

\begin{figure}[ht]
    \centering
    \includegraphics[width=\linewidth]{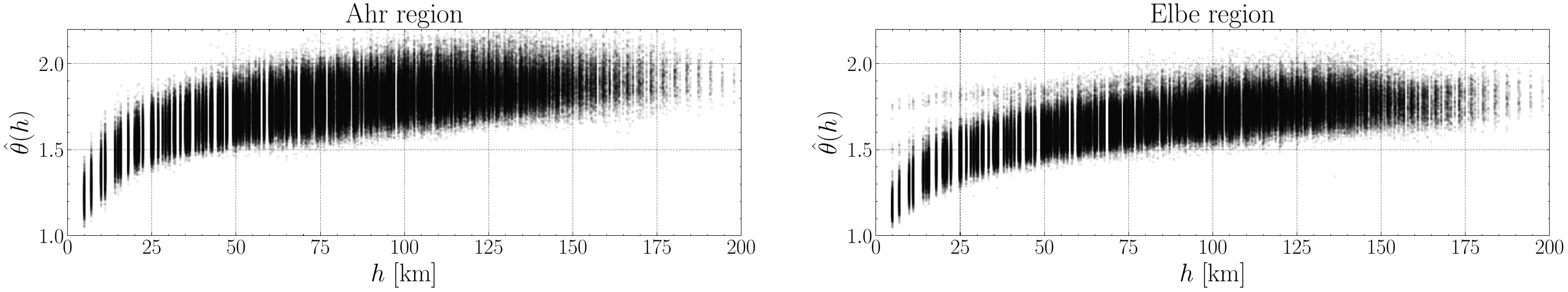}
    \caption{F-madogram estimates for both focus domains computed based on the entire available time period 1930-2025.}
    \label{fig:madogram_data}
\end{figure}
For the max-stable framework to apply, the data first needs to be transformed to unit Fr\'{e}chet margins. Here, we fit the marginals independently with a Generalized Extreme Value distribution (GEV), similar to \cite{Blanchet.2011}, since our focus lies on the spatial dependence rather than the marginal fit.
To further study the spatial dependence structure of the underlying process, we analyze the previously introduced F-madogram \eqref{eq:f_madogram}, which yields an estimator the extremal coefficient function $\theta(h)$. \autoref{fig:madogram_data} shows that near independence is reached at large distances for both domains, particularly for the Ahr region. This motivates the use of the Brown-Resnick model, whose extremal coefficient function can attain the upper bound of~2, unlike the Schlather model which is bounded below~2 and thus cannot represent independence. For further details on the spatial fit we refer to \autoref{app:gev}.
We rescale the data grid to units of $5\,\text{km}$ to match the simulation setting from Section~\ref{sec:simulation}, using the same support points for $\mathrm{EN}_\theta$.\footnote{This corresponds to a maximum spatial separation of $\|\bm{h}_{\max}\| \approx 42.5$.} The training simulations use $\lambda \in (0,50]$ and $\nu \in (0,2]$ to cover a broad range of dependence structures.

\begin{figure}[ht]
    \centering
    \includegraphics[width=\linewidth]{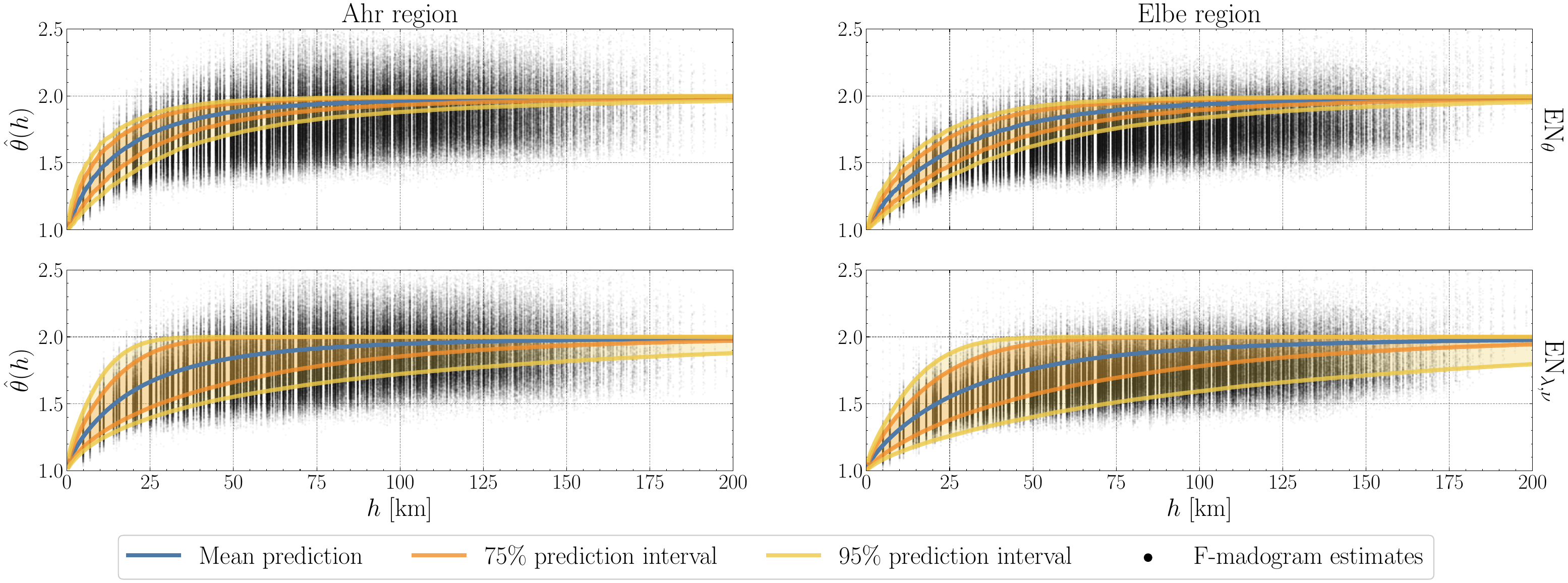}
    \caption{Estimated pairwise extremal coefficient function for the parametric ($\mathrm{EN}_{\lambda,\nu}$, lower) and direct non-parametric ($\mathrm{EN}_\theta$, upper) energy networks for both domains, averaged over 25 years (2000-2025). Black dots show the unbinned F-madogram estimates over the same period.}
    \label{fig:es_madogram_uncertainty}
\end{figure}
We first validate our approach by comparing the estimated spatial dependence $\hat{\theta}(h)$ of our model to the F-madogram estimations over the 25-year reference period that includes both flooding events. As illustrated  in \autoref{fig:es_madogram_uncertainty}, both models demonstrate an adequate fit to the empirical F-madograms, with $\mathrm{EN}_{\lambda,\nu}$ providing better calibrated prediction intervals, while $\mathrm{EN}_\theta$ appears somewhat overconfident.

\begin{figure}[ht]
    \centering
    \includegraphics[width=\linewidth]{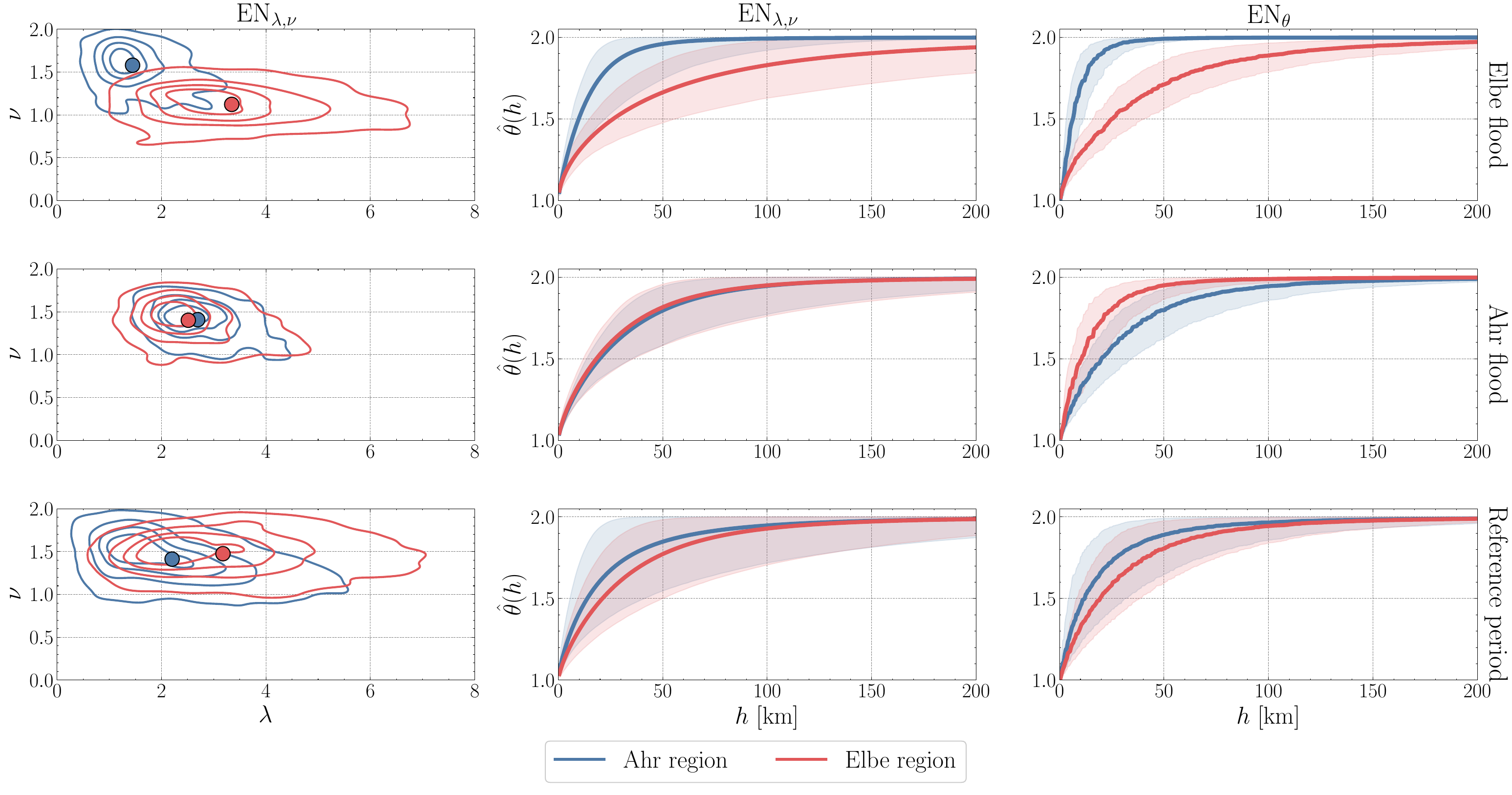}
    \caption{Estimation results for the Brown-Resnick parameters (left), the corresponding extremal coefficient function (middle), and the directly estimated extremal coefficient function (right). Shown are the mean estimates, parameter contour plots, and 90\% prediction intervals for $\hat{\theta}(h)$ for both the Ahr and Elbe regions during the respective event years (2002 and 2021) and the reference period 2003--2020.}
    \label{fig:application_estimates}
\end{figure}

Since our method produces estimates from a single process realization, we can directly compare the spatial dependence during individual extreme events. In particular, we study whether the spatial extremal dependence during the years 2002 and 2021, as well as the corresponding floods in the Elbe and Ahr regions, admits a different structure compared with an additional reference period, where no significant event occurred. \autoref{fig:application_estimates} shows estimates for the event years 2002 and~2021, alongside a reference period (2003--2020) without major flooding. For the reference period, the estimated dependence structures in both regions are similar and consistent across estimation approaches. The 2021 estimates closely match the reference period, indicating that the Ahr flood, while extreme in magnitude, does not exhibit an unusual spatial dependence structure.

For the Elbe flood in 2002, however, the estimations between both domains, as well as the reference period, differ significantly. The estimated smoothness parameter~$\nu$ is smaller, reflecting more fine-grained, localized precipitation clusters. Correspondingly, the extremal coefficient function for the Elbe region is significantly lower, indicating stronger spatial dependence at a given distance~$h$. This difference is visible in both estimation methods and significant at the 90\% prediction level. Compared to the reference period, the Elbe region in~2002 still shows elevated spatial dependence, particularly at larger distances. These results reveal that the two flooding events are substantially different in their spatial structure: the Elbe flood exhibits significantly stronger spatial dependence than both the Ahr flood and the reference period. We would conjecture that this difference might be due to the different geographical structure of the two areas. It seems that in plane areas flooding events are associated with increased spatial dependence.

\begin{figure}[ht]
    \centering
    \begin{subfigure}[b]{\textwidth}
    \centering
    \caption{Ahr region (2021)}
    \includegraphics[width=\linewidth]{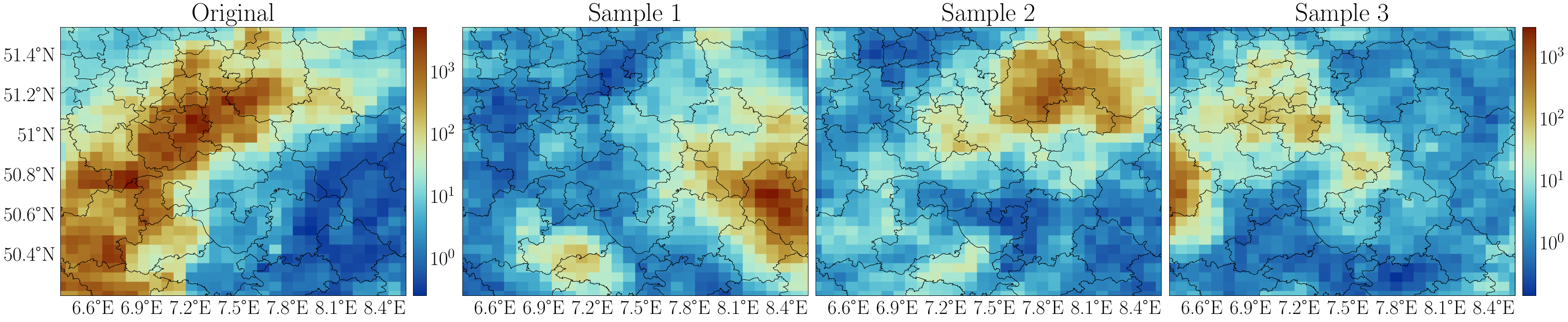}
    \end{subfigure}
    \begin{subfigure}[b]{\textwidth}
    \centering
    \caption{Elbe region (2002)}
    \includegraphics[width=\linewidth]{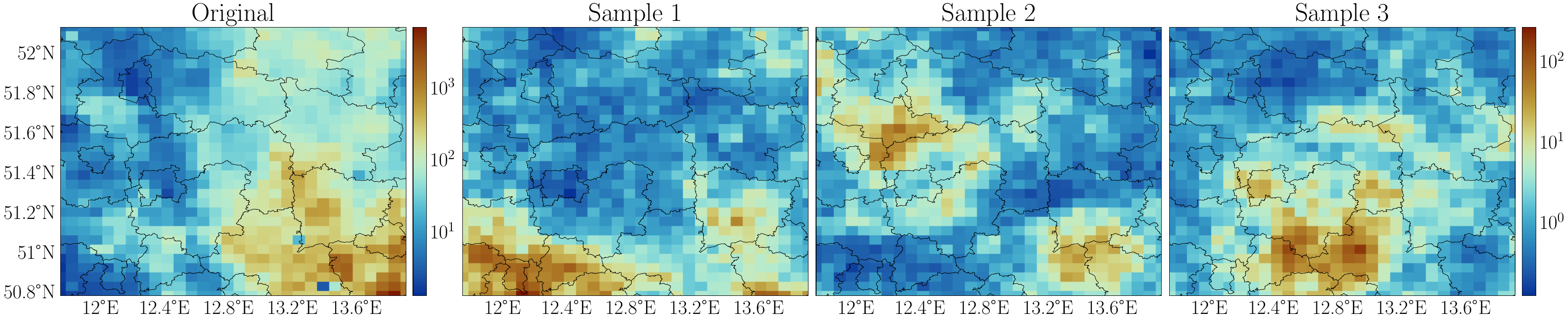}
    \end{subfigure}
    \caption{Ground truth precipitation maxima and samples from a Brown-Resnick process with parameters estimated by $\mathrm{EN}_{\lambda,\nu}$, transformed to the unit Fr\'{e}chet scale.}
    \label{fig:process_simulations}
\end{figure}

As a final validation, we compare observed extreme precipitation fields to Brown-Resnick processes simulated from the estimated (mean) parameters. \autoref{fig:process_simulations} shows that the Ahr flood produces larger, spatially coherent precipitation clusters, while the Elbe flood generates more localized extremes---consistent with the lower estimated smoothness~$\nu$ in \autoref{fig:application_estimates}. For both events, the simulated processes reproduce the spatial dependence of the original observations. The match is particularly accurate for the Elbe flood, while for the Ahr flood, the simulations admit slightly smaller clusters than the observed field, suggesting that the full spatial dependence may not be captured exactly. Note, however, that the magnitude of the events differs due to the estimated GEV parameters used for transformation, which cannot account for the marginal extremes of the flooding events. 

Overall, these results demonstrate the effectiveness of our approach for estimating the spatial dependence structure of extremal processes across distinct precipitation events. By producing full predictive distributions from single process realizations, the energy networks enable a direct comparison of individual extreme events, as well as underlying uncertainty\textemdash{}a task that is infeasible with classical estimators such as the F-madogram, which require averaging over many observations. Applied to the two most severe German flooding events of the last decades, our analysis reveals that while both events were extreme in magnitude, they differ fundamentally in their spatial dependence: the 2002 Elbe flood exhibits significantly stronger and more localized spatial dependence than the 2021 Ahr flood and the corresponding reference period, suggesting that the presented methodology can help characterize and distinguish different types of extreme precipitation settings.

%
%
\section{Discussion}
\label{sec:discussion}
We propose a general simulation-based approach to estimate and analyze the spatial dependence of precipitation maxima and their underlying uncertainty. Our method, based on generative neural networks, provides an estimate of the full predictive distribution of either the parameters of a suitable extremal model, such as max-stable processes, or of the pairwise extremal coefficient function and therefore of the corresponding spatial dependence. As a direct result, one obtains confidence estimates around the predicted parameters or the extremal coefficient function, enabling straightforward uncertainty quantification. By training the neural network on simulated extremal processes, we follow the general idea of likelihood-free inference and can adapt the framework to any model of extremal dependence.

We analyze the capabilities of our approach in a comprehensive simulation study, where the energy network demonstrates favorable performance across several metrics compared to different baselines. In addition, we validate robustness under several scenarios of model misspecification and show that the energy networks can be trained jointly across different max-stable models, highlighting their capacity to recover the true spatial dependencies even when the generating model is unknown.

Applied to summer precipitation maxima in Germany, we study two of the most severe flooding events of the last decades: the July 2021 Ahr valley flood in western Germany and the August 2002 Elbe flood in eastern Germany. Over the full 25-year reference period, the estimated extremal coefficient functions of both energy networks are well-calibrated against empirical F-madogram estimates, with the parametric network $\mathrm{EN}_{\lambda,\nu}$ producing better-calibrated prediction intervals than the nonparametric $\mathrm{EN}_\theta$. In particular, since our method yields full predictive distributions from single process realizations, we can compare the spatial dependence of individual extreme events---a task that is infeasible with classical estimators. This comparison reveals that the two floods differ fundamentally in their spatial structure: while the 2021 Ahr flood exhibits a spatial dependence pattern consistent with the non-event reference period (2003--2020), the 2002 Elbe flood shows significantly stronger spatial dependence, characterized by a lower estimated smoothness parameter and more localized, fine-grained precipitation clusters. This difference is significant at the 90\% prediction level and is consistent across both estimation approaches. These findings demonstrate that our method can distinguish different types of extreme precipitation regimes and may thus contribute to characterizing dangerous flood situations.

While our approach demonstrates good performance, it requires a substantial amount of simulations and training, which can impose a computational bottleneck. We therefore estimate the spatial dependence from a suitable parametric model rather then using nonparametric estimates 
Moreover, max-stable processes may only provide a worst-case bound on tail behavior, since by construction the pointwise block maxima aggregate over the timing and co-occurrence of extremes. In future work, the estimation framework could therefore benefit from incorporating more granular timing information as suggested by \cite{huser2025modeling}. While this work focuses on estimating spatial dependence under the assumption of a max-stable process, the general framework can be adapted to other suitable statistical models. An interesting research direction would be the application to multivariate threshold-exceedance approaches, such as the spatial conditional extremes model \citep{Wadsworth.2022, vandeskog2024fast}. Changing the underlying process only requires a change in the network architecture, whereas the general estimation strategy remains the same.

Another avenue for future work would be to further automate the estimation procedure, limiting the influence of tuning parameter choices. For instance, one could remove the need to specify a prior parameter range by implementing an iterative approach that automatically converges to the best estimation strategy. The predicted parameter samples could be used to simulate new processes iteratively until some stopping criterion is achieved, making the simulation process more efficient \citep{SainsburyDale.2022} and estimation more reliable. Providing theoretical conditions for such convergence would be of separate interest but would certainly require a different type of paper.

\clearpage
\bibliographystyle{ref.bst}
\bibliography{ref.bib}





\clearpage
\appendix

%
%

\section{Model architecture \& training details}
\label{app:architecture}
Both energy networks EN$_{\lambda,\nu}$ and EN$_\theta$ share the same architecture and hyperparameters, differing only in their final output layer. The input consists of spatial fields on a regular grid of $k=900$ locations on $\mathcal{D}=[0,30]^2$. The network comprises three blocks of convolutional and max-pooling layers, where the spatial output size decreases across blocks while the number of channels increases. Residual connections are added in the second and third block (compare \autoref{fig:model_architecture}). After the convolutional blocks, the representation is flattened and passed through fully connected layers.
 
All hidden layers use the ReLU activation function. For the output layer of EN$_{\lambda,\nu}$, we apply parameter-specific transformations: the smoothness parameter $\nu \in (0,2]$ is transformed to the unit interval and passed through a sigmoid activation, while the range parameter $\lambda > 0$ is log-transformed and uses a linear activation, following \cite{Lenzi.2023b}. For EN$_\theta$, the output values $\theta(h_1),\ldots,\theta(h_l)$ require no additional transformation and use a sigmoid activation scaled to the range $(1,2)$.
 
To produce $m=500$ posterior samples per prediction, the network samples from a latent space $\mathcal{N}(\boldsymbol{1}, \boldsymbol{I}_m)$ and multiplies the result onto a final linear layer. Both networks are trained by minimizing the energy score \eqref{eq:energy_score_estimation} using the RMSProp optimizer implemented in PyTorch with a learning rate of $7 \times 10^{-4}$ and a learning rate scheduler that terminates training when validation metrics cease to improve. The batch size is set to $100$.
 
The training set consists of $n=5000$ simulated spatial fields, with $20\%$ held out for validation. For EN$_\theta$, the pairwise extremal coefficient function is evaluated on a grid $h_\Delta$ with spacing $\Delta=0.1$ up to an upper bound of $h = \sqrt{30^2+30^2}$. To increase training efficiency, we apply data augmentation via $180^\circ$ rotation (probability $0.3$) and vertical and horizontal flips (probability $0.2$), exploiting the stationarity and isotropy of the Brown-Resnick and Schlather models, which ensure that these transformations preserve the dependence structure. Hyperparameters were selected based on preliminary experiments; systematic hyperparameter tuning is left for future work.
 
All experiments were conducted on a workstation equipped with an Intel Xeon E5-2680 2.50\,GHz CPU (40 cores) and an NVIDIA GeForce RTX 2080 GPU (8\,GB RAM).

\FloatBarrier
\section{Runtime analysis}
\label{app:runtime}
We analyze the computational cost of the neural network-based methods (CNN, EN$_{\lambda, \nu}$, EN$_{\theta}$, Log$_{\lambda, \nu}$) and compare them to the likelihood-based approaches PL and ABC. Table~\ref{tab:runtime_analysis} summarizes the final test loss, training time, and inference time for all methods using the full training dataset of size $n=5000$ for training and the test set of size $n=250$.

\begin{table}[ht]
\centering
\caption{Final test loss, as well as time for training ($n=5000$) and inference over the whole test set ($n=250$) for all methods.}
\label{tab:runtime_analysis}
\begin{tabular}{lllllll}
\toprule
 & CNN & EN$_{\lambda, \nu}$ & EN$_{\theta}$ & Log$_{\lambda, \nu}$ & PL & ABC \\
\midrule
Training Time& 127.98s (20.46) & 159.62s (17.86) & 103.66s (7.43) & 240.55s (51.65) & - & -\\
Inference Time & 0.03s (0.00) & 0.04s (0.00) & 0.03s (0.00) & 0.03s (0.00) & $\sim$1h & $\sim$30h \\
\bottomrule
\end{tabular}
\end{table}

All three neural network architectures train in under four minutes on the full dataset, with EN$_{\theta}$ being the fastest at $103.66$s, followed by CNN at $127.98$s and EN$_{\lambda, \nu}$ at $159.62$s. However, when considering the standard deviation across multiple runs, these three methods essentially require the same runtime. The Log$_{\lambda, \nu}$ method takes slightly longer with $240.55$s, likely due to the computation of the loss function and unstable convergence.

Once trained, all three models produce predictions in approximately $0.03$--$0.04$s, enabling near-instantaneous simulation-based inference. In contrast, the PL method requires roughly one hour and ABC approximately $30$ hours per inference run, making them impractical for settings where repeated or real-time predictions are needed. It should be noted, however, that similar to the baseline methods, the neural network methods require simulated processes for training, which can be substantial depending on the complexity of the simulator. Once this initial cost is amortized, inference on new observations is practically instantaneous.

\begin{figure}[ht]
    \centering
    \includegraphics[width=\linewidth]{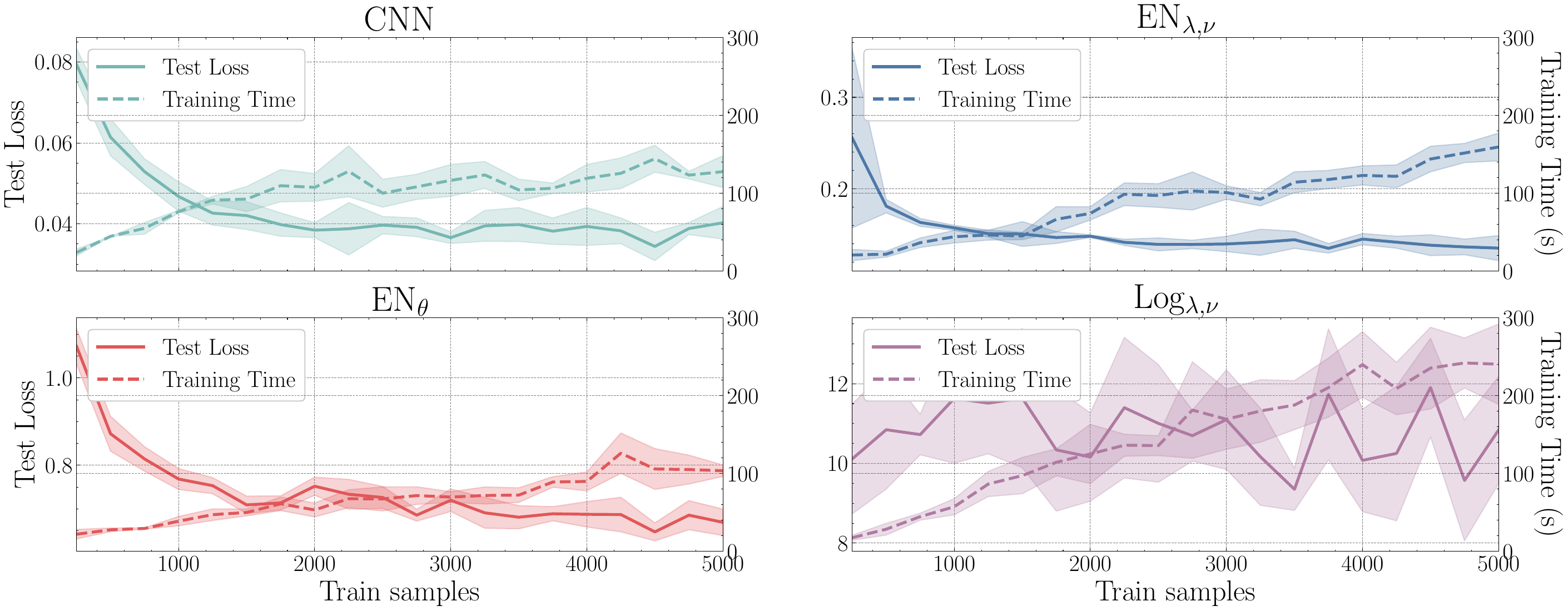}
    \caption{Test loss and training time as a function of the number of training samples. Results are averaged over five runs, with shaded areas indicating one standard deviation.}
    \label{fig:runtime_analysis}
\end{figure}

Figure~\ref{fig:runtime_analysis} shows how test loss and training time scale with the number of training samples. Training time increases approximately linearly with dataset size for all three architectures. Notably, even at smaller training set sizes, the neural network methods achieve competitive test losses, with diminishing improvements beyond roughly $n=3000$ samples. EN$_{\lambda,\nu}$ and CNN exhibit a steeper initial decrease in test loss compared to EN$_{\theta}$, suggesting that the energy-based formulation with learned bandwidth benefits more strongly from additional training data in the small-sample regime. For the Log$_{\lambda, \nu}$ method, training is less stable.

\section{Log-score multivariate normal comparison}
\label{app:log_score}
Here, we compare our model to a parametric predictive model and analyze the differences between the energy score and the log-score. Since the log-score requires a predictive density, it cannot be combined with our sampling-based approach. To provide a comparison, we instead consider a predictive multivariate Gaussian with diagonal covariance matrix for the parameter estimation task. In particular, we assume that
\[
\mathbb{Q}_{\phi}(\cdot \mid \boldsymbol{Z}^{[i]}) = \mathcal{N}\left(\cdot; (\mu_\lambda, \mu_\nu)^\top, \mathrm{diag}(\sigma_\lambda^2, \sigma_\nu^2) \mid \boldsymbol{Z}^{[i]} \right),
\]
i.e. each parameter, $\lambda, \nu$ follows an i.i.d. Gaussian distribution. Now, we can use the log-score, which admits a closed-form expression for the Gaussian, to optimize the neural network with respect to the predicted parameters. In addition, we also analyze this setup when using the energy score as a loss function, which works the same way as before, by generating samples from the distribution $\left(\hat{\boldsymbol{\gamma}}_j(i)\right)_{j=1}^m \sim \mathbb{Q}_{\phi}(\cdot\mid\boldsymbol{Z}^{[i]})$. Finally, we also include the comparison of our nonparametric model, optimized using the energy score. We refer to the three models as EN$_{\lambda, \nu}$ for the nonparametric energy score model and MVN-EN$_{\lambda,\nu}$, Log$_{\lambda,\nu}$ for the multivariate normal energy and log-score models, respectively. 

\begin{figure}[ht]
    \centering
    \begin{subfigure}[b]{\textwidth}
    \centering
    \caption{Brown-Resnick}
    \includegraphics[width=\linewidth]{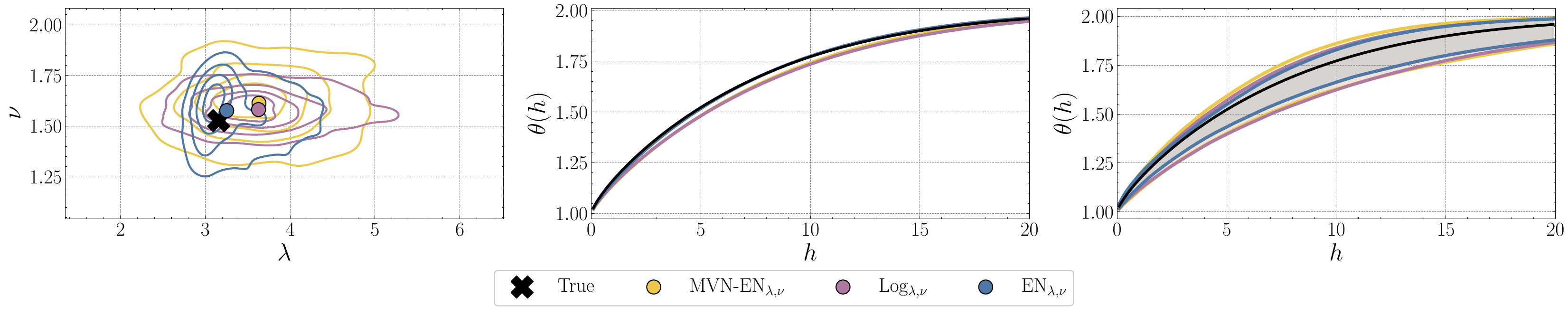}
    \end{subfigure}
        \begin{subfigure}[b]{\textwidth}
    \centering
    \caption{Powered exponential}
    \includegraphics[width=\linewidth]{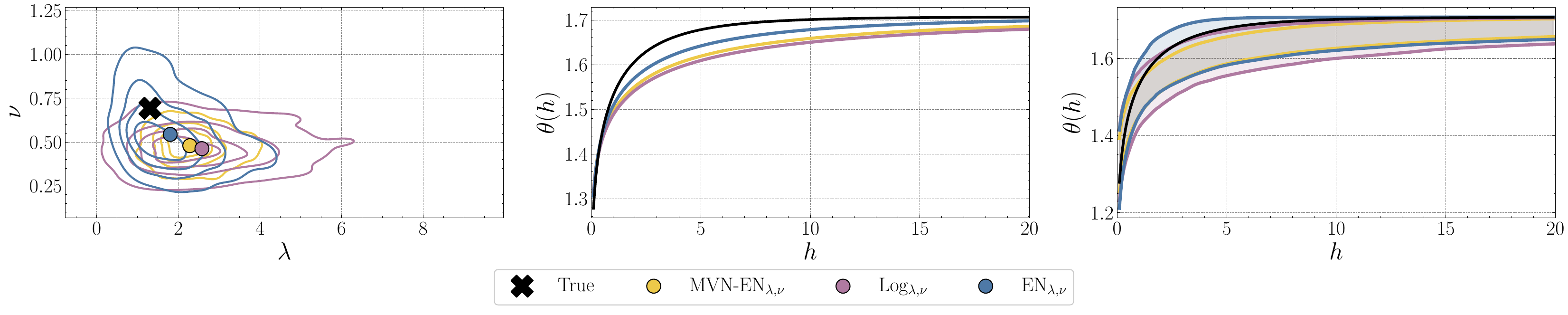}
    \end{subfigure}
    \caption{The figure visualizes the different estimation methods for the Brown-Resnick (a) and Powered exponential (b) model using a randomly drawn test samples. Further specifications are as in \autoref{fig:results_single}.}
    \label{fig:log_score_comparison_normal}
\end{figure}

\begin{figure}[ht]
    \centering
    \begin{subfigure}[b]{\textwidth}
    \centering
    \caption{Misspecified parameter range}
    \includegraphics[width=\linewidth]{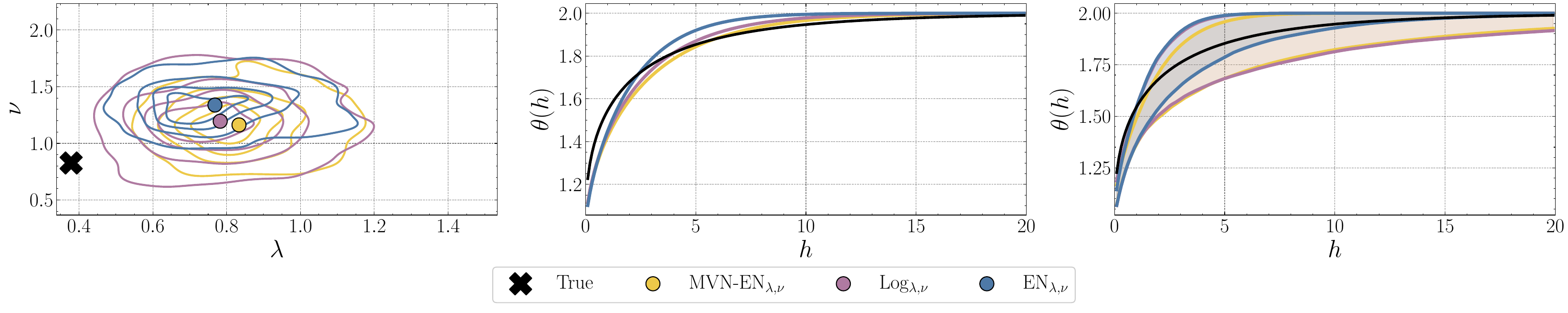}
    \end{subfigure}
    \begin{subfigure}[b]{\textwidth}
    \centering
    \caption{Misspecified correlation function}
    \includegraphics[width=\linewidth]{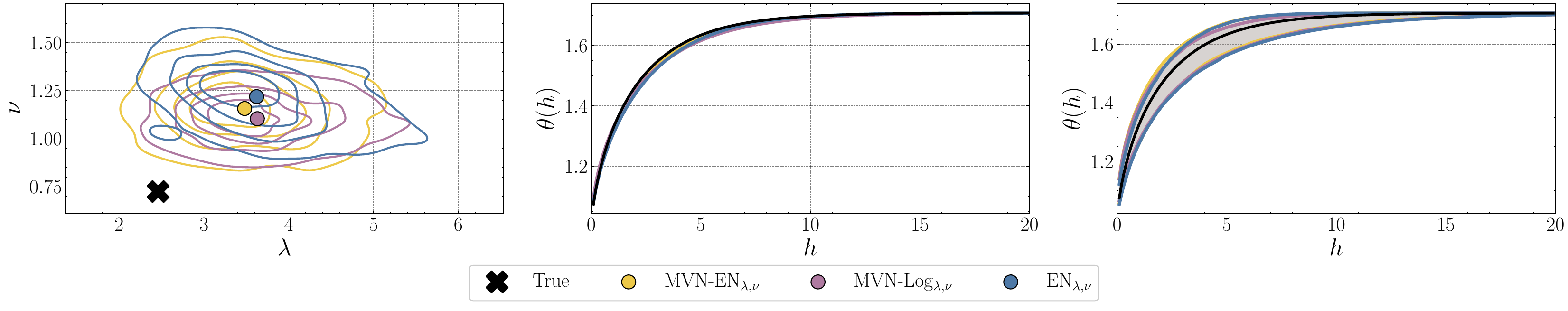}
    \end{subfigure}
    \begin{subfigure}[b]{\textwidth}
    \centering
    \caption{Misspecified model}
    \includegraphics[width=\linewidth]{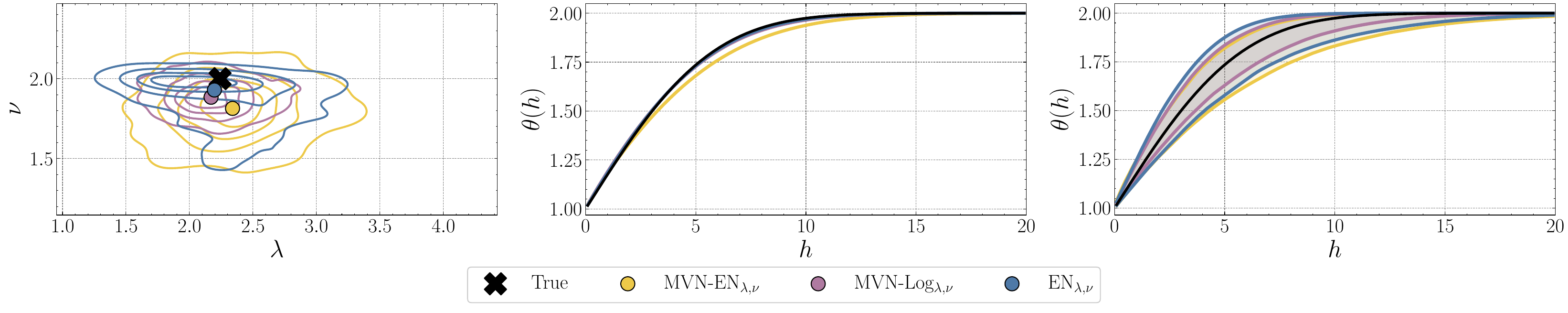}
    \end{subfigure}
    \caption{The figure visualizes the different estimation methods for the different robustness scenarios using a randomly drawn test samples. Further specifications are as in \autoref{fig:results_single}.}
    \label{fig:log_score_comparison_robust}
\end{figure}

\autoref{tab:log_score_results} shows the evaluation metrics for all three networks across the different max-stable models and robustness scenarios, while \autoref{fig:log_score_comparison_normal} and \autoref{fig:log_score_comparison_robust} show selected visualizations. It is noteworthy that the Log$_{\lambda,\nu}$ model performs very well for the regular Brown-Resnick and Powered exponential scenarios, better than the networks based on the energy-score. However, this is to be expected, as when the model is correctly specified, minimizing the log-score leads to the estimator with the lowest possible variance, as it approaches the Cramér-Rao bound and therefore leads to better performance as compared to the MVN-EN$_{\lambda,\nu}$ model. 
However, for the scenarios with model misspecification, the models based on the energy score, especially the nonparametric EN$_{\lambda,\nu}$ model, lead to better performance, most notably for the probabilistic metrics. 
So while the log-score works as an alternative scoring rule to optimize a neural network that parametrizes a predictive Gaussian, our energy network is more robust to misspecification, which is highly relevant in application settings, where the true model is unavailable.

\begin{table}[ht]
\centering
\caption[Robustness split results]{Table a) shows averages of probabilistic performance scores (see Section \ref{subsec:eva}) for the obtained empirical distribution of the posterior across the different datasets and evaluation metrics. In Table b) we depict results for pointwise errors of the respective pointwise estimates. All metrics are negatively oriented, with the best model highlighted in bold and respective standard deviations are given in brackets. Table c) shows the empirical coverage of the estimated intervals for different levels, where the optimal value is the corresponding interval width.}
\label{tab:log_score_results}
\footnotesize

\caption*{\textbf{(a) Pointwise evaluation of average estimators}}
\begin{tabular}{|c|c|c|c|c|}
\hline
\textbf{Scenario} & \textbf{Estimator} & \textbf{MSE\textsubscript{$\lambda$}} & \textbf{MSE\textsubscript{$\nu$}} & \textbf{MSE\textsubscript{$\theta$}} \\
\hline

\multirow{3}{*}{\textbf{Brown-Resnick}} 
& EN$_{\lambda,\nu}$ & \textbf{0.34 (0.63)} & \textbf{0.01 (0.02)} & \textbf{0.15 (0.26)} \\
\cline{2-5}
& MVN-EN$_{\lambda,\nu}$ & 0.48 (1.06) & 0.02 (0.03) & 0.18 (0.32) \\
\cline{2-5}
& Log$_{\lambda,\nu}$ & 0.43 (0.88) & 0.02 (0.03) & 0.17 (0.29) \\
\hline
\hline

\multirow{3}{*}{\textbf{Powexp}} 
& EN$_{\lambda,\nu}$ & 0.52 (1.19) & 0.03 (0.05) & \textbf{0.01 (0.02)} \\
\cline{2-5}
& MVN-EN$_{\lambda,\nu}$ & 0.55 (1.23) & \textbf{0.02 (0.04)} & \textbf{0.01 (0.02) }\\
\cline{2-5}
& Log$_{\lambda,\nu}$ & \textbf{0.47 (1.07)} & \textbf{0.02 (0.04)} & \textbf{0.01 (0.02)} \\
\hline
\hline

\multirow{3}{*}{\makecell{\textbf{Misspecified} \\ \textbf{parameter range}}}
& EN$_{\lambda,\nu}$ &\textbf{ 9.17 (10.06)} & \textbf{0.04 (0.06)} & \textbf{0.64 (0.63)} \\
\cline{2-5}
& MVN-EN$_{\lambda,\nu}$ & 9.66 (10.42) & 0.05 (0.07) & 0.68 (0.63) \\
\cline{2-5}
& Log$_{\lambda,\nu}$ & 9.77 (10.35) & \textbf{0.04 (0.09)} & 0.72 (0.67) \\
\hline
\hline

\multirow{3}{*}{\makecell{\textbf{Misspecified} \\ \textbf{correlation function}}} 
& EN$_{\lambda,\nu}$  & \textbf{1.08 (1.26)} & 0.23 (0.19) & 0.05 (0.07) \\
\cline{2-5}
& MVN-EN$_{\lambda,\nu}$ & 1.18 (1.49) & \textbf{0.19 (0.17)} & 0.05 (0.08) \\
\cline{2-5}
& Log$_{\lambda,\nu}$& 1.32 (1.53) & \textbf{0.19 (0.15)} & \textbf{0.04 (0.07)} \\
\hline
\hline

\multirow{3}{*}{\makecell{\textbf{Misspecified} \\ \textbf{model}}} 
& EN$_{\lambda,\nu}$ & \textbf{0.04 (0.08)} & \textbf{0.01 (0.04)} & \textbf{0.02 (0.05)} \\
\cline{2-5}
& MVN-EN$_{\lambda,\nu}$ & 0.19 (0.21) & 0.03 (0.04) & 0.07 (0.07) \\
\cline{2-5}
& Log$_{\lambda,\nu}$ & 0.05 (0.07) & \textbf{0.01 (0.02)} & \textbf{0.02 (0.02)} \\
\hline
\end{tabular}

\caption*{\textbf{(b) Probabilistic evaluation of the posterior distribution estimates}}

\begin{tabular}{|c|c|c|c|c|c|c|}
\hline
\textbf{Scenario} & \textbf{Estimator} & \textbf{IS\textsubscript{0.05,$\lambda$}} & \textbf{IS\textsubscript{0.05,$\nu$}} & \textbf{ES} & \textbf{IIS\textsubscript{0.05}} & \textbf{ES\textsubscript{$\theta$}} \\
\hline

\multirow{3}{*}{\textbf{Brown-Resnick}}
& EN$_{\lambda,\nu}$ & 2.91 (4.80) & \textbf{0.53 (0.59)} & \textbf{0.33 (0.31)} & 12.89 (21.18) & \textbf{1.88 (0.85)} \\
\cline{2-7}
& MVN-EN$_{\lambda,\nu}$  & 3.64 (8.11) & 0.61 (0.92) & 0.38 (0.36) & 13.53 (21.50) & \textbf{1.88 (0.83) }\\
\cline{2-7}
& Log$_{\lambda,\nu}$ & \textbf{2.89 (5.41)} & \textbf{0.53 (0.77)} & 0.35 (0.32) & \textbf{12.03 (19.90)} & \textbf{1.88 (0.84)} \\
\hline
\hline

\multirow{3}{*}{\textbf{Powexp}}
& EN$_{\lambda,\nu}$ & 3.50 (7.69) & 0.75 (1.04) & 0.40 (0.38) & 2.53 (5.22) & \textbf{0.45 (0.17)} \\
\cline{2-7}
& MVN-EN$_{\lambda,\nu}$  & 4.32 (10.04) & 0.75 (0.92) & 0.41 (0.41) & 2.86 (5.52) & \textbf{0.45 (0.17)} \\
\cline{2-7}
& Log$_{\lambda,\nu}$ & \textbf{3.23 (6.97)} & \textbf{0.61 (0.86)} & \textbf{0.36 (0.33)} & \textbf{2.31 (6.01)} & \textbf{0.45 (0.16)} \\
\hline
\hline

\multirow{3}{*}{\makecell{\textbf{Misspecified} \\ \textbf{parameter range}}}
& EN$_{\lambda,\nu}$ & 71.17 (62.09) & \textbf{1.74 (2.28) }& \textbf{2.28 (1.66)} & 89.24 (79.35) & \textbf{2.83 (1.23)} \\
\cline{2-7}
& MVN-EN$_{\lambda,\nu}$ & \textbf{60.63 (56.60)} & 2.18 (3.02) & 2.30 (1.63) & \textbf{81.32 (75.51)} & 2.85 (1.25) \\
\cline{2-7}
& Log$_{\lambda,\nu}$ & 63.61 (59.62) & 2.52 (3.47) & 2.34 (1.65) & 99.67 (90.11) & 2.89 (1.25) \\
\hline
\hline

\multirow{3}{*}{\makecell{\textbf{Misspecified} \\ \textbf{correlation function}}} 
& EN$_{\lambda,\nu}$& \textbf{12.10 (14.42)} & 8.27 (6.91) & \textbf{0.82 (0.40)} & \textbf{10.25 (15.45)} & 0.70 (0.47) \\
\cline{2-7}
& MVN-EN$_{\lambda,\nu}$ & 14.72 (17.65) & \textbf{8.10 (6.84)} & 0.84 (0.49) & 14.25 (21.67) & 0.70 (0.48) \\
\cline{2-7}
& Log$_{\lambda,\nu}$ & 18.87 (20.86) & 10.40 (6.96) & 0.93 (0.51) & 14.01 (21.08) & \textbf{0.68 (0.45)} \\
\hline
\hline

\multirow{3}{*}{\makecell{\textbf{Misspecified} \\ \textbf{model}}} 
& EN$_{\lambda,\nu}$  & \textbf{0.94 (0.71)} & 1.45 (1.24) & \textbf{0.13 (0.08)} & \textbf{2.13 (2.37)} & \textbf{0.58 (0.35)} \\
\cline{2-7}
& MVN-EN$_{\lambda,\nu}$ & 1.49 (1.38) & 0.65 (0.93) & 0.28 (0.15) & 4.24 (4.79) & 0.72 (0.42) \\
\cline{2-7}
& Log$_{\lambda,\nu}$ & 1.13 (1.39) & \textbf{0.32 (0.91)} & 0.15 (0.10) & 2.89 (3.43) & 0.60 (0.38) \\
\hline
\end{tabular}

\caption*{\textbf{(c) Coverage of the posterior distribution estimates}}

\begin{tabular}{|c|c|c|c|c|c|c|c|c|c|c|}
\hline
\multirow{2}{*}{\textbf{Model}} &\multirow{2}{*}{\textbf{Estimator}}  & \multicolumn{3}{c|}{$95\%$} & \multicolumn{3}{c|}{$75\%$}& \multicolumn{3}{c|}{$50\%$} \\
\cline{3-11}
 &  & $\lambda$ & $\nu$ & $\theta$ & $\lambda$ & $\nu$ & $\theta$ & $\lambda$ & $\nu$ & $\theta$\\
\hline
\multirow{3}{*}{\textbf{Brown-Resnick}} 
& EN$_{\lambda,\nu}$ & 0.81 & 0.91 & 0.84 & 0.57 & 0.67 & 0.63 & \textbf{0.42} & 0.44 & 0.40 \\
\cline{2-11}
& MVN-EN$_{\lambda,\nu}$ & 0.87 & \textbf{0.95} & \textbf{0.88} & 0.64 & \textbf{0.77} & \textbf{0.70} & \textbf{0.42} & \textbf{0.49} & \textbf{0.47} \\
\cline{2-11}
& Log$_{\lambda,\nu}$ & \textbf{0.93} & 0.94 & 0.87 & \textbf{0.69} & 0.69 & 0.65 & 0.40 & \textbf{0.49} & 0.40 \\

\hline
\hline
\multirow{3}{*}{\textbf{Powexp}} 
& EN$_{\lambda,\nu}$& 0.90 & 0.89 & \textbf{0.89} & \textbf{0.69} & 0.66 & \textbf{0.69} & \textbf{0.47 }& 0.44 & \textbf{0.51} \\
\cline{2-11}
& MVN-EN$_{\lambda,\nu}$  & 0.84 & 0.90 & 0.86 & 0.63 & \textbf{0.69} & 0.65 & 0.40 & \textbf{0.46 }& 0.46 \\
\cline{2-11}
& Log$_{\lambda,\nu}$& \textbf{0.92} & \textbf{0.91} & 0.88 & 0.68 & 0.68 & 0.67 & 0.43 & 0.44 & 0.46 \\

\hline
\hline
\multirow{3}{*}{\makecell{\textbf{Misspecified} \\ \textbf{parameter range}}} 
& EN$_{\lambda,\nu}$ & 0.18 & \textbf{0.56} & 0.27 & 0.10 & \textbf{0.38} & \textbf{0.17} & 0.06 & \textbf{0.21} & \textbf{0.11} \\
\cline{2-11}
& MVN-EN$_{\lambda,\nu}$ & 0.20 & 0.47 & \textbf{0.29} & \textbf{0.11} & 0.26 & 0.13 & \textbf{0.07} & 0.15 & 0.08 \\
\cline{2-11}
& Log$_{\lambda,\nu}$ & \textbf{0.23} & 0.41 & 0.25 & 0.09 & 0.24 & 0.12 & \textbf{0.07} & 0.10 & 0.08 \\

\hline
\hline
\multirow{3}{*}{\makecell{\textbf{Misspecified} \\ \textbf{correlation function}}} 
& EN$_{\lambda,\nu}$ & \textbf{0.54} & \textbf{0.28} & \textbf{0.47} & \textbf{0.38} & 0.13 & 0.29 & \textbf{0.22} & 0.07 & 0.19 \\
\cline{2-11}
& MVN-EN$_{\lambda,\nu}$ & 0.48 & 0.25 & \textbf{0.47} & 0.35 & \textbf{0.14 }& \textbf{0.35} & \textbf{0.22} & \textbf{0.09 }& \textbf{0.21 }\\
\cline{2-11}
& Log$_{\lambda,\nu}$ & 0.40 & 0.12 & 0.42 & 0.25 & 0.08 & 0.32 & 0.14 & 0.04 & \textbf{0.21} \\

\hline
\hline
\multirow{3}{*}{\makecell{\textbf{Misspecified} \\ \textbf{model}}} 
& EN$_{\lambda,\nu}$ & \textbf{0.95} & 0.00 & \textbf{0.85} & \textbf{0.79} & 0.00 & \textbf{0.58} & \textbf{0.53} & 0.00 & \textbf{0.40} \\
\cline{2-11}
& MVN-EN$_{\lambda,\nu}$ & 0.93 & \textbf{0.92} & 0.78 & 0.49 & \textbf{0.42} & 0.28 & 0.21 & \textbf{0.02} & 0.14 \\
\cline{2-11}
& Log$_{\lambda,\nu}$ & 0.76 & 0.87 & 0.63 & 0.44 & 0.07 & 0.35 & 0.27 & 0.00 & 0.20 \\
\hline
\end{tabular}

\end{table}

\FloatBarrier
\section{Additional figures}
\label{app:additional_visualizations}

\begin{figure}[ht]
\centering
\begin{subfigure}[b]{0.9\textwidth}
    \caption{Brown-Resnick}
    \includegraphics[width = \textwidth]{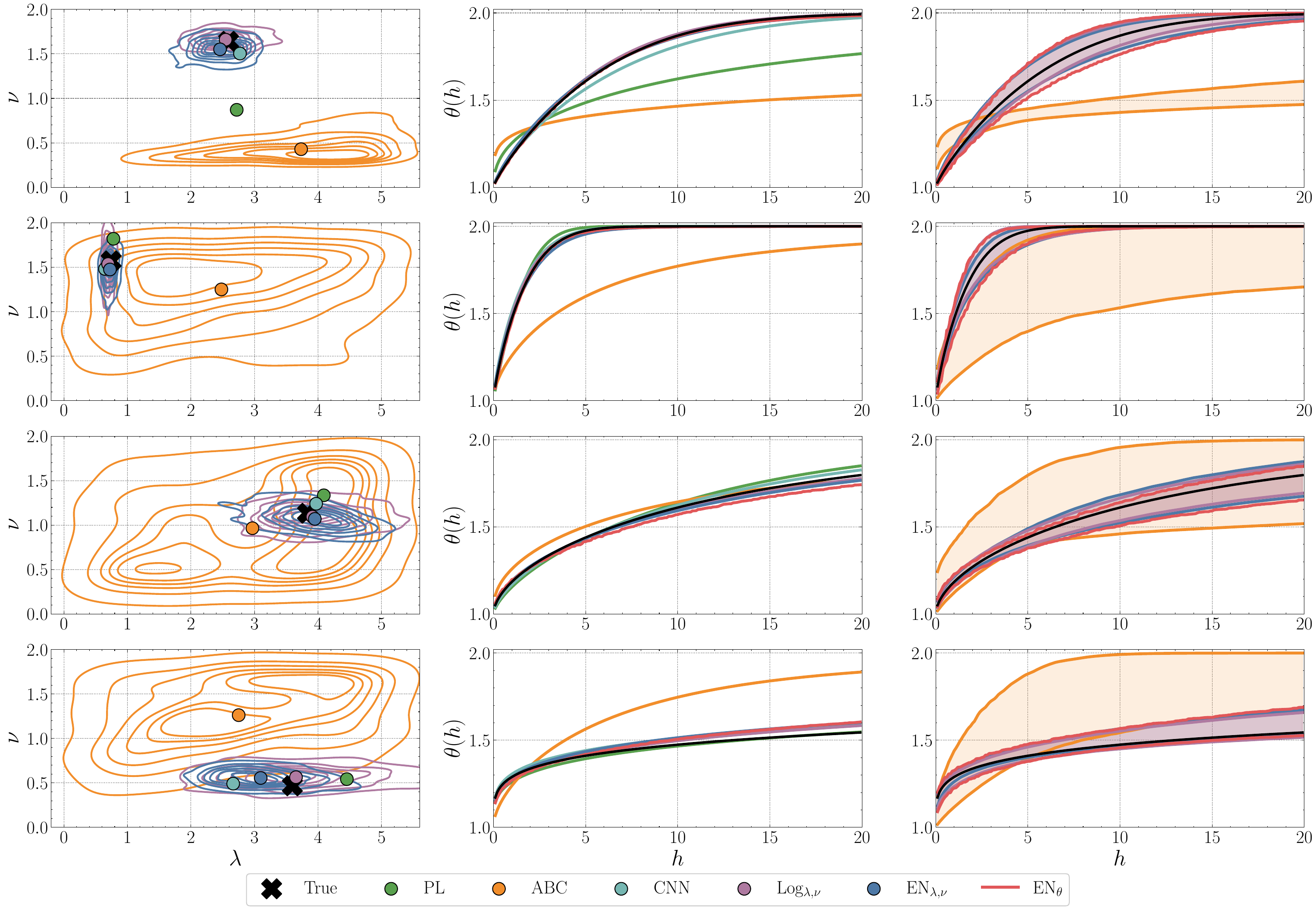}
\end{subfigure}
\begin{subfigure}[b]{0.9\textwidth}
    \caption{Powered exponential}
   \includegraphics[width = \textwidth]{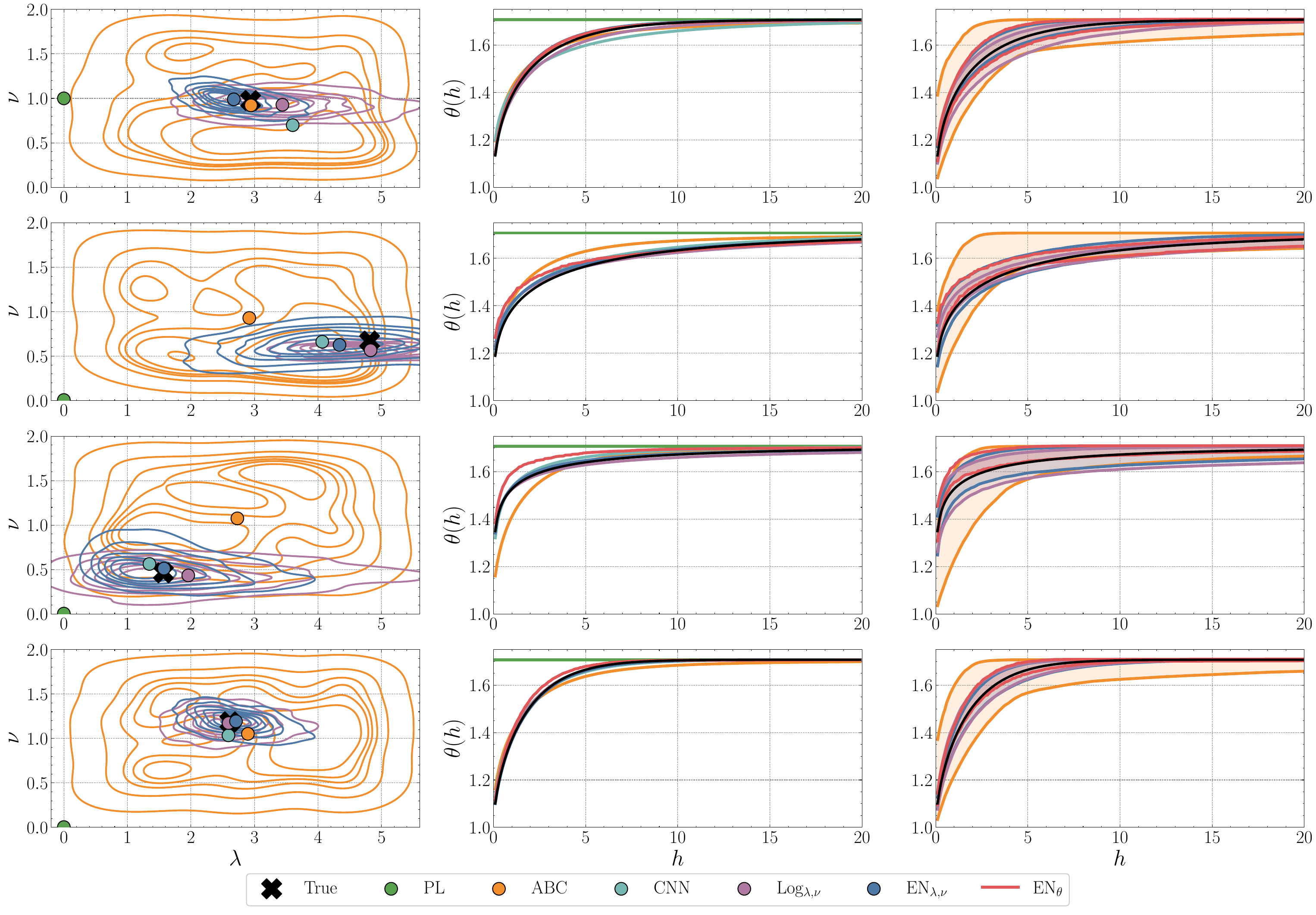}
\end{subfigure}
\caption{The figure visualizes the different estimation methods for the max-stable models using four randomly drawn test samples. Further specifications are as in \autoref{fig:results_single}.}
\label{fig:normal_results_multiple}
\end{figure}

\begin{figure}[ht]
\centering
\begin{subfigure}[b]{0.9\textwidth}
    \caption{Misspecified parameter range}
    \includegraphics[width = \textwidth]{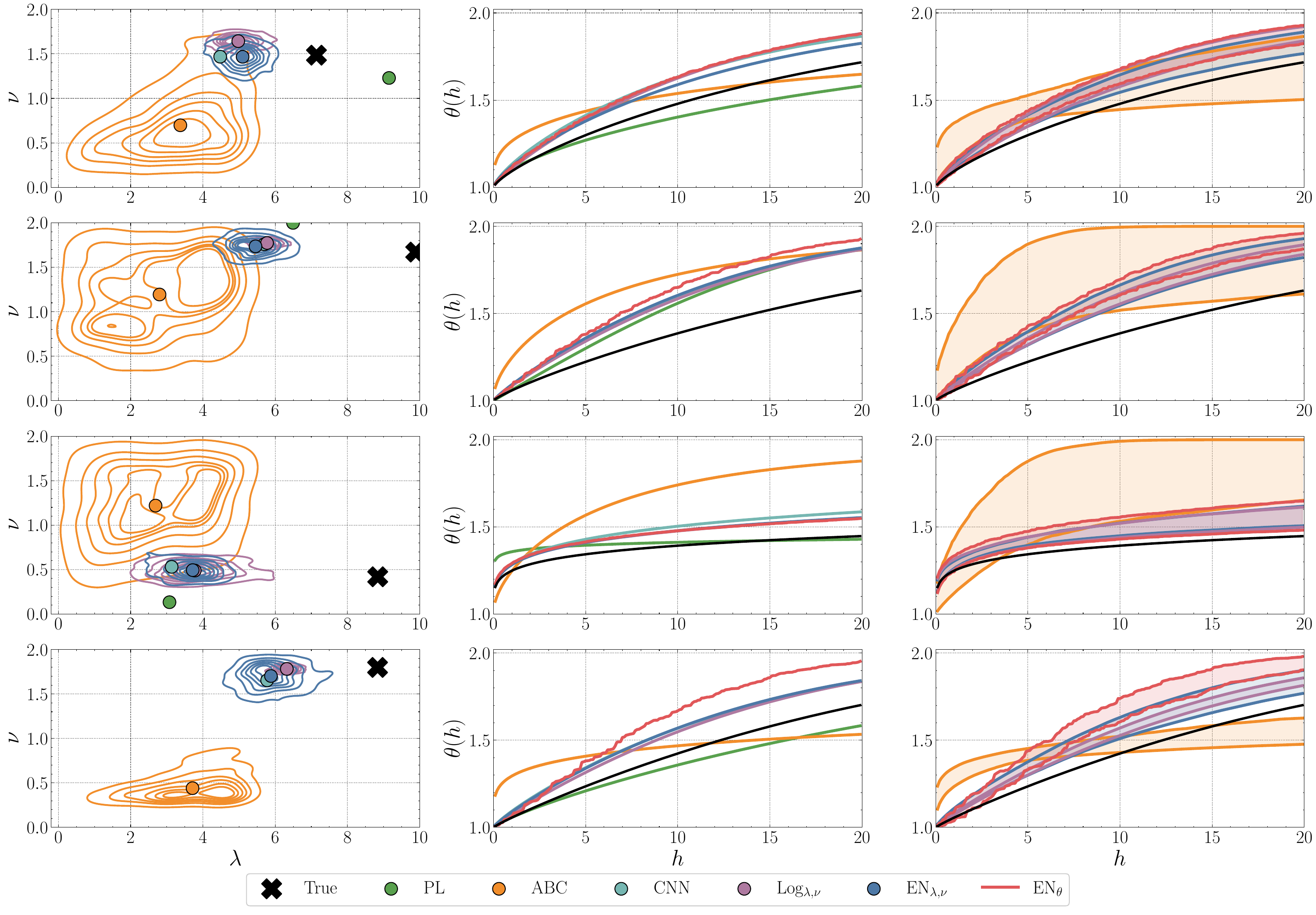}
\end{subfigure}
\begin{subfigure}[b]{0.9\textwidth}
    \caption{Misspecified correlation function}
   \includegraphics[width = \textwidth]{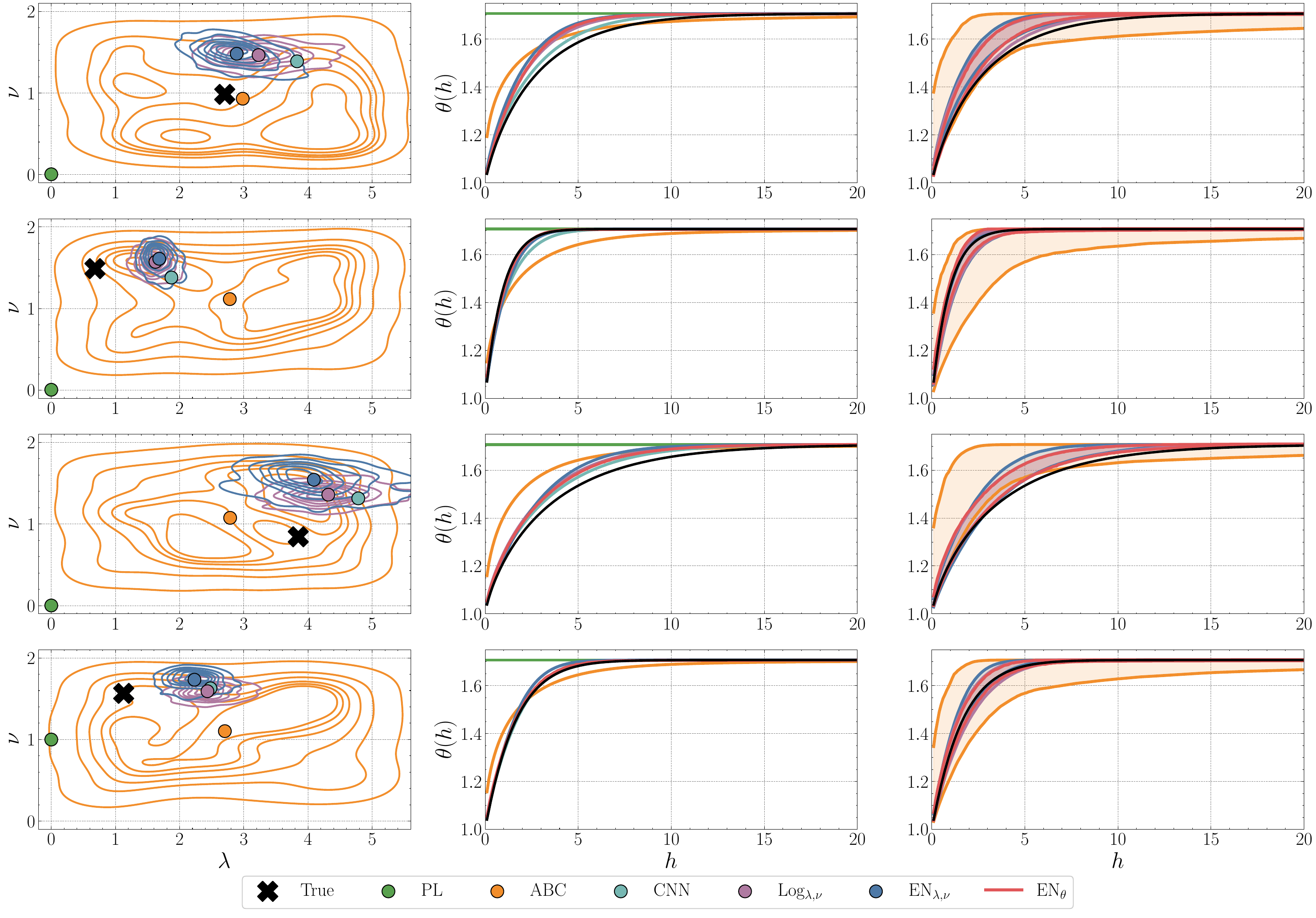}
\end{subfigure}
\caption{The figure visualizes the different estimation methods for the first two robustness scenarios using four randomly drawn test samples. Further specifications are as in \autoref{fig:results_single}.}
\label{fig:robust_results_multiple}
\end{figure}

\begin{figure}[ht]
\centering
    \includegraphics[width = \textwidth]{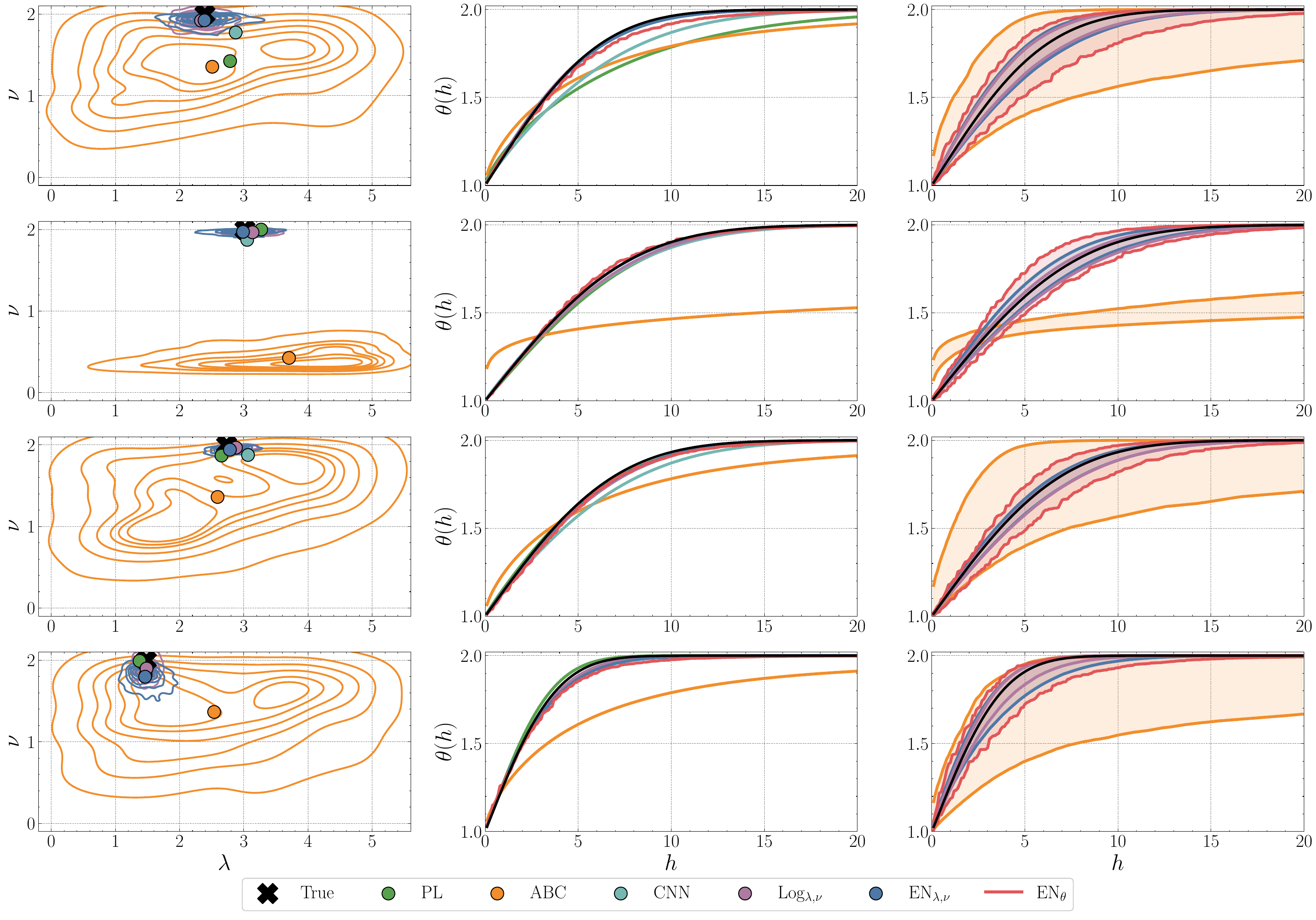}
\caption{The figure visualizes the different estimation methods for the third robustness scenario using four randomly drawn test samples. Further specifications are as in \autoref{fig:results_single}.}
\label{fig:robust_results_multiple_smith}
\end{figure}

%
%
\clearpage
\section{Retaining monotonicity}
\label{app:sorting}

\begin{figure}[ht]
    \centering
    \begin{subfigure}[b]{\textwidth}
        \caption{Mean prediction}
        \includegraphics[width=\linewidth]{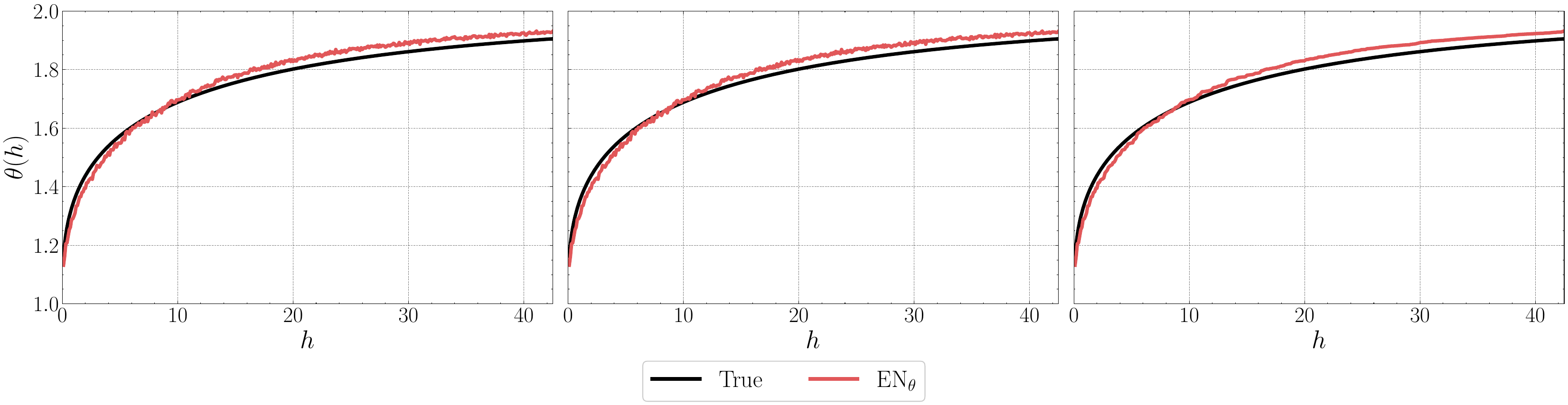}
    \end{subfigure}
    \begin{subfigure}[b]{\textwidth}
        \caption{Quantile prediction}
        \includegraphics[width=\linewidth]{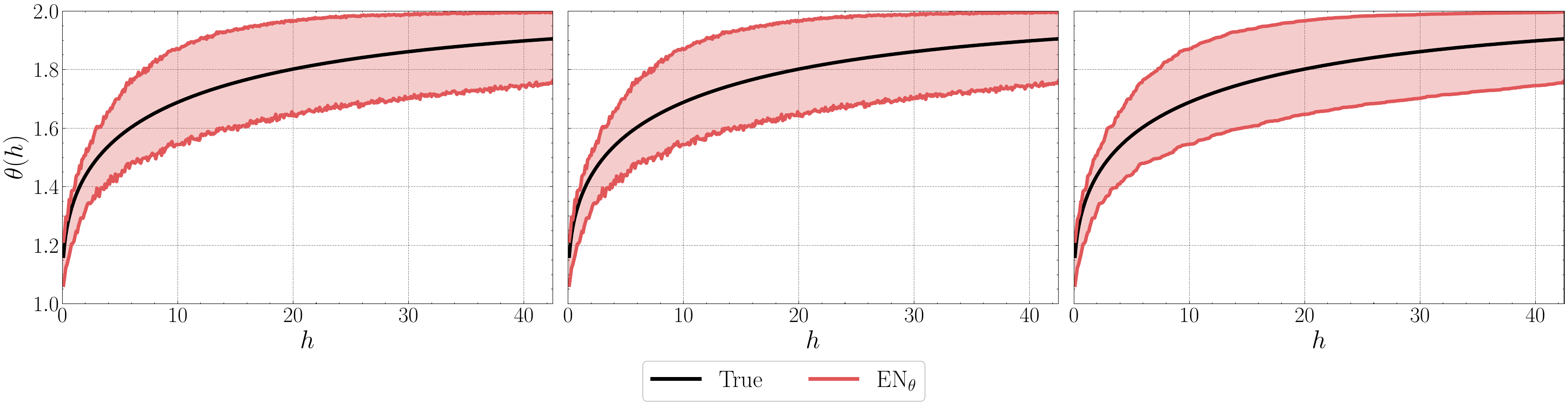}
    \end{subfigure}
    \caption{Estimates of the pairwise extremal coefficient function using EN$_\theta$ with different ways of sorting. On the left, the values $\hat{\theta}(h_i),\, i=1,\ldots,n$ are not sorted, only the functionals are evaluated on the given sample. In the middle, the values are sorted first and the functional is evaluated afterwards, while on the right, the functional is first evaluated and the values are sorted in ascending order afterwards.}
    \label{fig:sorting}
\end{figure}

For the direct estimation of $\boldsymbol{\theta}(h_{\Delta})$ with EN$_\theta$ it can no longer be guaranteed that $\hat{\theta}_r(h_i)\leq \hat{\theta}_r(h_j)$ for $i\leq j,\, \forall r=1,\ldots,m$. The problem can be solved by sorting the estimated values in ascending order. However, initially it is not clear whether to first sort each sample or whether to first calculate the functional of interest. As visualized in \autoref{fig:sorting} we found that first evaluating the functional of interest, i.e., the mean or a certain quantile and sorting the values afterward, for example, $\overline{\hat{\theta}(h_1)},\ldots,\overline{\hat{\theta}(h_n)}$, leads to the best prediction, as it retains monotonicity and leads to more smooth predictions of the extremal coefficient function.

%
%
\clearpage
\section{GEV fit \& Spatial dependence}
\label{app:gev}
\begin{figure}[ht]
    \centering
    \begin{subfigure}{\linewidth}
        \caption{Ahr region}
            \includegraphics[width = \linewidth]{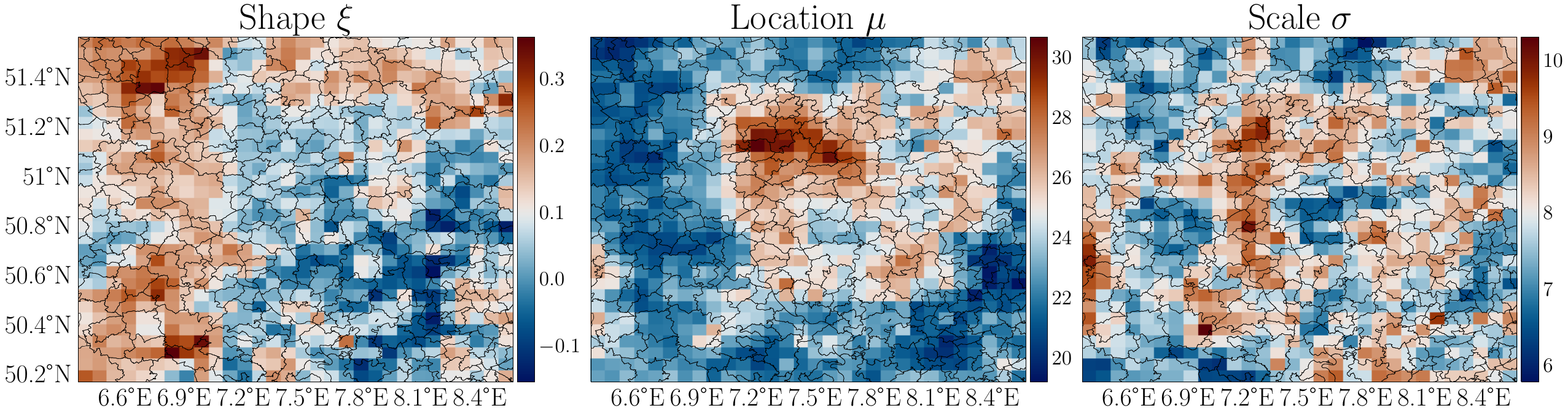}
    \end{subfigure}
    \begin{subfigure}{\linewidth}
        \caption{Elbe region}
            \includegraphics[width = \linewidth]{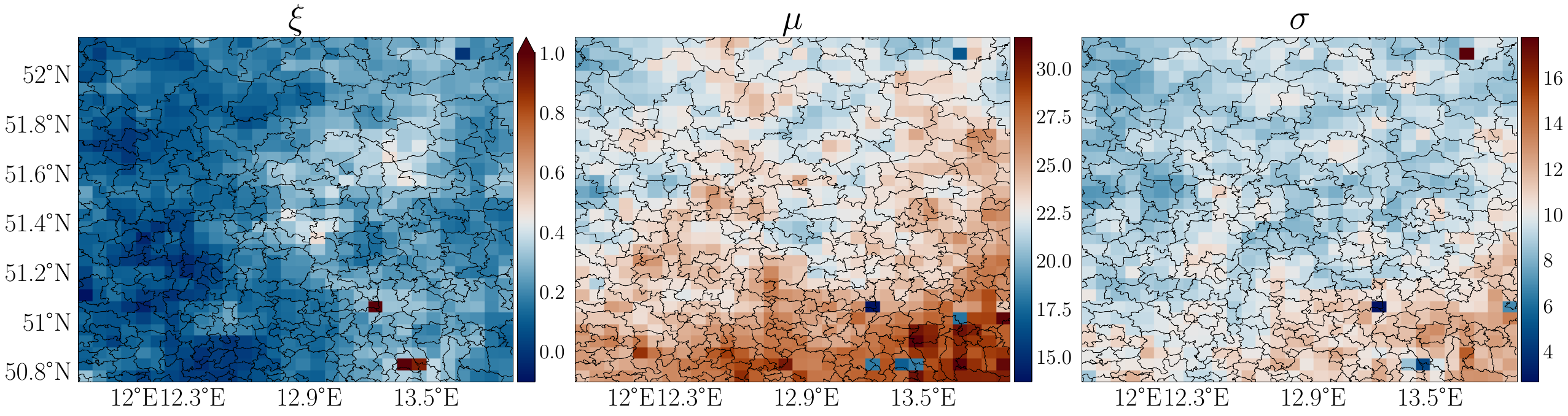}
    \end{subfigure}
    \caption{Visualizations of the marginally fitted GEV parameters $(\xi, \mu, \sigma)$ over the two different domains.}
    \label{fig:gev_surfaces}
\end{figure}

\begin{figure}[h]
    \centering
    \includegraphics[width=\linewidth]{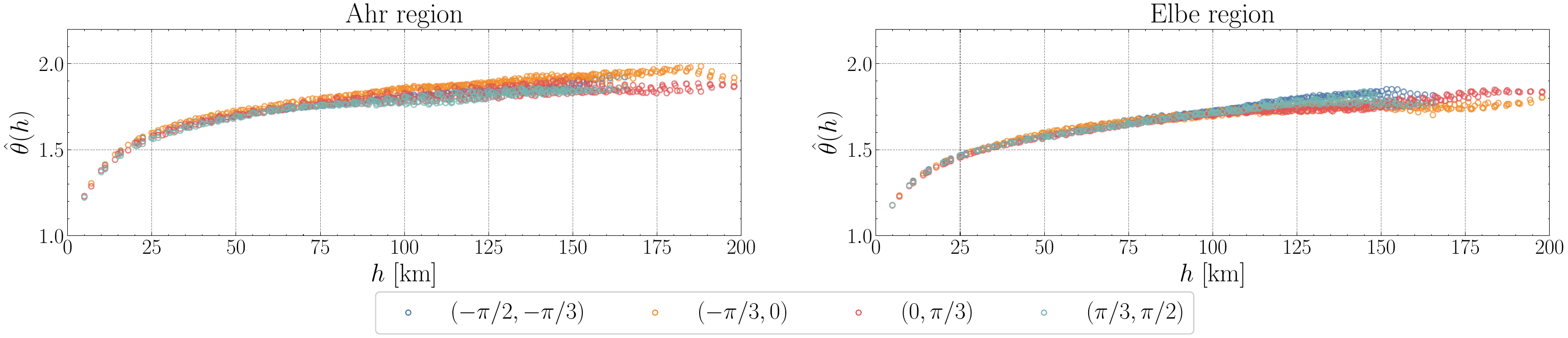}
    \caption{F-madogram estimates (bin means), aggregated by four different directions for both different domains based on the entire time period 1930-2015.}
    \label{fig:madogram_isotropic}
\end{figure}

In order to transform the data to the unit Fréchet scale, we fit via Maximum-Likelihood the parameters of the GEV distribution \citep{coles2001introduction}

\begin{equation}
G(z)
=
\exp\left(
-
\left\{
1+\xi(x)\frac{z-\mu(x)}{\sigma(x)}
\right\}_{+}^{-1/\xi(x)}
\right),
\label{eq:gev}
\end{equation}
with $u_{+} = \max(u,0)$ and $\mu(x) \in \mathbb{R}$, $\sigma(x)>0$, and $\xi(x) \in \mathbb{R}$ denoting the location, scale, and shape parameters, respectively. For each gridpoint, we fit $\xi, \mu, \sigma$ over the full available time period  1930-2025. While other approaches, for example, modeling the parameters based on covariates such as latitude or longitude, have been considered \citep{Padoan.2010}, we choose a simple marginal modeling approach, similar to \citep{Blanchet.2011} in order to rather focus on spatial dependence. Despite their simplicity, however, the obtained GEV-estimates yield a good marginal fit.\footnote{The multiple respective diagnostic plots are available upon request.}
The parameters are estimated using the \emph{genextreme} distribution class from \emph{scipy} \citep{Virtanen.2020}. The final estimated parameters at each grid point for both domains are visualized in \autoref{fig:gev_surfaces}.
Note that the shape parameter is (generally) estimated as positive, indicating that the data can best be described using a Fréchet distribution. This fits naturally into the precipitation scenario, as the Fréchet distribution has a left endpoint, corresponding to the nonnegativity of precipitation values. Using the obtained estimates, the observed precipitation fields are transformed to unit Fréchet margins.
Given the transformed surfaces, we further check the extremal dependence for isotropy, again using the F-madogram estimator. \autoref{fig:madogram_isotropic} shows the binned means of $\hat{\theta}(h)$ in different directional sectors, allowing to analyze eventual anisotropy of the underlying process. For both regions, isotropy seems justified for distances up to 150km, and mild anisotropy is visible only for very large distances, for which less data is available, justifying the use of the isotropic max-stable process as an underlying model.

\end{document}